\PassOptionsToPackage{numbers,sort&compress}{natbib}
\documentclass[onecolumn]{fairmeta}
\usepackage[most]{tcolorbox}

\usepackage{amsmath}
\usepackage{amssymb}
\usepackage[utf8]{inputenc} 
\usepackage[T1]{fontenc}    
\usepackage{hyperref}       
\usepackage{url}            
\usepackage{booktabs}       
\usepackage{amsfonts}       
\usepackage{nicefrac}       
\usepackage{graphicx}
\usepackage{enumitem}

\usepackage{algorithm}
\usepackage{algpseudocode}
\usepackage{multirow}
\usepackage{microtype}      
\usepackage{xcolor}         
\newcommand{\m}[1]{{A2A}}
\newcommand{\p}[1]{}

\usepackage{wrapfig}

\title{Feedback World Model Enables Precise Guidance of Diffusion Policy}

\author[1,*]{\href{https://morpheus-an.github.io/}{\textcolor{black}{Tuo An}}}
\author[1,*]{\href{https://jiajindou.github.io/}{\textcolor{black}{Jindou Jia}}}
\author[1]{\href{https://reagan1311.github.io/}{\textcolor{black}{Gen Li}}}
\author[1]{Jingliang Li}
\author[1]{\href{https://chuhaozhou99.github.io/Chuhao-Zhou/}{\textcolor{black}{Chuhao Zhou}}}
\author[1]{Pengfei Liu}
\author[1]{Bofan Lyu}
\author[1]{Jiaqi Bai}
\author[1]{Xinying Guo}
\author[1]{Geng Li}
\author[1]{\href{https://marsyang.site/}{\textcolor{black}{Jianfei Yang}}}

\affiliation[1]{MARS Lab, Nanyang Technological University}

\setcitestyle{numbers,square,comma}
\newif\ifCRdraft
\CRdrafttrue
\definecolor{CRorange}{RGB}{160,75,0}

\newcommand{\CRplaceholder}[3]{%
  \fbox{\begin{minipage}[c][#2][c]{0.94\linewidth}
    \centering\color{CRorange}\textbf{Figure pending: #1}\par\medskip
    \small #3
  \end{minipage}}}
\newcommand{\CRfigure}[4]{%
  \IfFileExists{Figs/camera_ready/#1}{\includegraphics[width=\linewidth]{Figs/camera_ready/#1}}%
  {\CRplaceholder{#2}{#3}{#4}}}

\algrenewcommand\algorithmiccomment[1]{\hfill{\color{gray!90}\(\triangleright\)~\textit{#1}}}
\newenvironment{ack}{\section*{Acknowledgments and Disclosure of Funding}}{}

\usepackage{needspace}
\usepackage{placeins}
\AtBeginDocument{%
  \setcounter{topnumber}{4}%
  \setcounter{bottomnumber}{3}%
  \setcounter{totalnumber}{6}%

}

\newcommand{\MarsLabAppendixLayout}{%
  \pretocmd{\section}{\ifnum\value{section}=13\Needspace{26\baselineskip}\fi}{}{}%
}

\abstract{
World models aim to improve robotic decision making by predicting the consequences of actions.
However, in practice, their predictions often become unreliable once the robot encounters states outside the training distribution, limiting their effectiveness at deployment.
We observe that execution itself provides a natural but underutilized signal: after each action, the robot directly observes the true next state, revealing the mismatch between predicted and actual outcomes.
Building on this insight, we propose \textbf{feedback world model}, an inference-time mechanism that closes the loop between prediction and observation at inference time.
Instead of treating the world model as a static open-loop predictor, our method maintains a lightweight feedback state that is updated online to iteratively correct future predictions, compensating for model errors using real-time observations without additional training data or parameter updates.
We show that this process can be interpreted as a latent-space observer and admits convergence guarantees under bounded, asymptotically convergent dynamics residuals and a stable feedback gain.
We further introduce action-aware guidance to better translate corrected predictions into control by emphasizing action-controllable components while suppressing irrelevant variations.
Experiments on LIBERO-Plus, Robomimic, and real-world manipulation tasks demonstrate that our method substantially improves both prediction accuracy and policy performance under distribution shift, reducing world model prediction error by up to 76.4\%.
The same feedback mechanism improves the predictions of the pretrained foundation model Fast-WAM-Joint without updating its parameters, demonstrating its plug-and-play applicability.
These results show that incorporating real-time feedback at inference time provides a simple yet powerful alternative to static world modeling.

}

\correspondence{\email{antu0001@e.ntu.edu.sg}, \email{jianfei.yang@ntu.edu.sg}}
\metadata[Project site]{\url{https://ntumars.github.io/Feedback_World_Model/}}
\contribution[*]{Equal contribution}

\begin{document}
\maketitle

\section{Introduction}
\label{sec:introduction}
World models, which predict how environment states evolve under actions, have emerged as a promising tool for improving the robustness and generalization of robotic policies~\citep{vla-jepa,fast-wam,dreamzero, world-guidance, world-model-for-robot-learning,WM_Survey}. Existing approaches primarily incorporate world models in two ways. One line integrates predicted dynamics, preferences, or rewards into policy training, using them as auxiliary supervision or optimization signals~\citep{vla-jepa,fast-wam,atomvla,world-vla-loop}.
The other leverages world models at inference time to evaluate or plan over candidate actions, directly influencing decision making during execution~\citep{latent-policy-barrier,progressVLA}. Across both paradigms, the key idea is to move beyond pure behavior cloning by grounding policy learning and execution in predicted environment dynamics, encouraging actions that lead to plausible and task-consistent future states~\citep{eval_robot_in_worldmodel,latent-policy-barrier}.

However, these benefits become difficult to realize when policies are deployed beyond the training distribution.
In robotic manipulation tasks, even small variations in initial pose, object configuration, or visual observation can shift the system into out-of-distribution (OOD) regimes.
Under such shifts, the reliability of the world model becomes critical: inaccurate predictions can directly mislead both policy learning and test-time guidance~\citep{eval_robot_in_worldmodel}. 
A common strategy to improve robustness is to scale up data or model capacity, for example through large pretrained video models~\citep{dreamzero} or additional real-world rollouts for finetuning~\citep{latent-policy-barrier,world-vla-loop}.
While effective, this approach substantially increases training cost and is often impractical in data-limited robotic settings~\citep{dreamzero,fast-wam}.
Notably, deployment itself provides an underexplored signal: after each action, the robot observes the true next state, which directly exposes the mismatch between predicted and actual transitions.
This real-time feedback offers a natural opportunity to improve prediction reliability without additional data collection or model scaling.

For guidance based on a frozen world model, conditioning each prediction on the latest observation leaves a further opportunity: retaining past prediction residuals to correct subsequent predictions.
Such online correction can help the predictor track deployment conditions as an episode unfolds.
This motivates our central question:
\textbf{can a world model actively exploit real-time feedback during inference to maintain reliable prediction under distribution shift, without requiring additional data or larger models?}


To this end, we propose a feedback world model that leverages real-time observations during policy inference to correct future predictions online. 
Instead of operating as a static open-loop predictor, our model maintains a lightweight feedback state that is updated after each environment interaction.
The observed mismatch between predicted and observed transitions is used to iteratively correct subsequent predictions, mitigating error accumulation without requiring additional training data or parameter updates.
This mechanism can be interpreted as a latent-space observer with a convergence guarantee under the residual and stability assumptions.
Furthermore, we introduce action-aware guidance strategy for diffusion policy inference.
Rather than uniformly comparing predicted observations, we emphasize components that are more directly influenced by the robot’s actions, such as end-effector motion and object pose.
This focuses guidance on controllable, task-relevant changes while reducing interference from action-irrelevant variations.

We evaluate our method on LIBERO-Plus~\citep{libero-plus}, Robomimic~\citep{robomimic}, and real-world manipulation tasks, with distribution shifts spanning robot initial states, lighting, backgrounds, object layouts, and sensor noise.
The experiments examine both world model prediction accuracy and policy performance, and include feedback correction of a frozen pretrained Fast-WAM-Joint model across the full LIBERO-10 suite.
Without additional training data or parameter updates, feedback reduces world-model prediction error by up to $76.4\%$.
When combined with action-aware guidance, it improves policy robustness in simulation and on real robots, matching or exceeding the base policy's success rate in 18 of 20 task--perturbation settings on LIBERO-Plus.

To summarize, our contributions are threefold: (1) We propose a feedback world model that incorporates real-time observations as online corrective signals, correcting frozen predictions with low incremental feedback cost and a guarantee under explicit residual and stability conditions.
(2) We introduce an action-aware guidance strategy that emphasizes controllable, action-relevant components to improve policy guidance.
(3) Experiments in simulation and on real robots demonstrate improved prediction accuracy and policy performance under distribution shift, with further gains when transferring our feedback mechanism to the pretrained Fast-WAM-Joint model.

\section{Related work}
\label{sec:related_work}
\paragraph{Efficient and accurate world models.}
Recent work on robotic world models has improved model quality along three directions: stronger latent representations, more accurate predictive dynamics, and more efficient architectures. Representation learning spans geometric and foundation-model approaches in robotics~\citep{deng2025best3dscenerepresentation}. VLA-JEPA uses latent predictive learning~\citep{vla-jepa}, while DINO-WM learns dynamics over pretrained visual features~\citep{dido-wm}. Another improves predictive accuracy through iterative or closed-loop refinement, where world models are adapted using rollout data or coupled with downstream learning~\citep{world-vla-loop,fowm}. In parallel, efficiency-oriented designs such as Fast-WAM~\citep{fast-wam}, DreamZero~\citep{dreamzero}, and DDP-WM~\citep{ddp-wm} reduce the cost of world or world-action modeling through latent-space prediction, architectural simplification, or faster inference.

\paragraph{Latent filtering and belief models.}
Alongside improvements in representations and dynamics, latent filtering reconciles model predictions with new observations.
DVBF~\citep{dvbf} learns state-space models through variational inference. PlaNet and Dreamer~\citep{planet,dreamer,dreamerv2} jointly learn recurrent latent dynamics and observation-conditioned inference, using observations to update their latent state during execution.
Our analytic feedback update adds residual correction to an already trained predictor without training an additional inference network or changing model parameters.
This enables direct reuse of pretrained world models, while their corrected predictions provide more reliable guidance for diffusion policies.

\paragraph{Inference-time policy steering.}
Recent efforts steer pretrained policies at inference time for improved robustness. Since diffusion policies generate actions through iterative denoising~\citep{diffsuion-policy}, they naturally allow external objectives to be injected during sampling. Value-Guided Policy Steering~\citep{value-guidance} re-ranks sampled actions using a learned value function, DynaGuide~\citep{dynaguide} injects guidance from an external latent dynamics model, and latent policy barrier (LPB)~\citep{latent-policy-barrier} uses future latent predictions relative to an expert manifold. More structured guidance signals include contact-guided generation in hierarchical diffusion policy~\citep{hierarchical-guidance} and task-progress guidance in ProgressVLA~\citep{progressVLA}.

\section{Preliminaries}
\label{sec:preliminaries}
\subsection{Action-conditioned world model for robot learning}
In robot learning, world models are commonly used as action-conditioned
predictive models that estimate how the robot-environment state evolves under
candidate actions. 
In this work, we consider prediction in a latent state space. Let $O_t$ denote
the observation context available at time $t$, which may include the current
observation as well as a finite history of previous observations.
We encode this context into a latent state
\begin{equation}
    z_t = \psi(O_t),
\end{equation}
where $\psi$ is the context encoder. For notational brevity, we denote a
candidate future action sequence from time $t$ as $A_t$. The latent world model
$f_\theta$ then predicts the next latent state as
\begin{equation}
\label{world_model}
    \hat{z}_{t+1}=f_\theta(z_t,A_t).
\end{equation}
This formulation captures action-conditioned latent dynamics and enables
rollout-based reasoning about the consequences of candidate robot actions~\citep{planet,dreamer,dreamerv2}.

\subsection{Diffusion policy}

Generative policies model conditional action distributions through diffusion~\citep{diffsuion-policy} or flow matching~\citep{jia2026action}. We use a diffusion policy that generates action sequences through iterative denoising. The policy learns a time-dependent score function that estimates the gradient of the log-probability of actions conditioned on the observation and optional language instruction~\citep{generative-score-songyang,songscore-sde}.

\paragraph{Score function learning.}
A diffusion policy $\pi_{\phi}$ parameterized by $\phi$ learns the score function:
\begin{equation}
    s_{\phi}({A}_{t}^{(\tau)}, O_t, \ell, \tau) \approx \nabla_{{A}_{t}^{(\tau)}} \log p_{\tau}({A}_{t}^{(\tau)} \mid O_t, \ell),
\end{equation}
where ${A}_{t}^{(\tau)}$ is the noisy action at diffusion step $\tau$, and $\ell$ denotes an optional language instruction. In practice, this is implemented via a denoising objective that trains the model to predict the score (or equivalently, the injected noise) at each diffusion step.

\paragraph{Guided denoising.}
Since the diffusion policy explicitly models the score function, additional objectives can be naturally incorporated at inference time via score guidance~\citep{diffusion_beat_gans}. Given a task-oriented guidance term $g({A}_{t}^{(\tau)}, O_t)$, the guided score is:
\begin{equation} \label{guidance}
    \tilde{s}({A}_{t}^{(\tau)}, O_t, \ell, \tau) = s_{\phi}({A}_{t}^{(\tau)}, O_t, \ell, \tau) + \lambda \, g({A}_{t}^{(\tau)}, O_t),
\end{equation}
where $\lambda$ controls the strength of guidance. This formulation enables integrating auxiliary signals, such as those derived from world models, into the action generation process~\citep{plan-with-diffusion, latent-policy-barrier}.

\begin{figure}
    \centering    \includegraphics[width=1.0\linewidth]{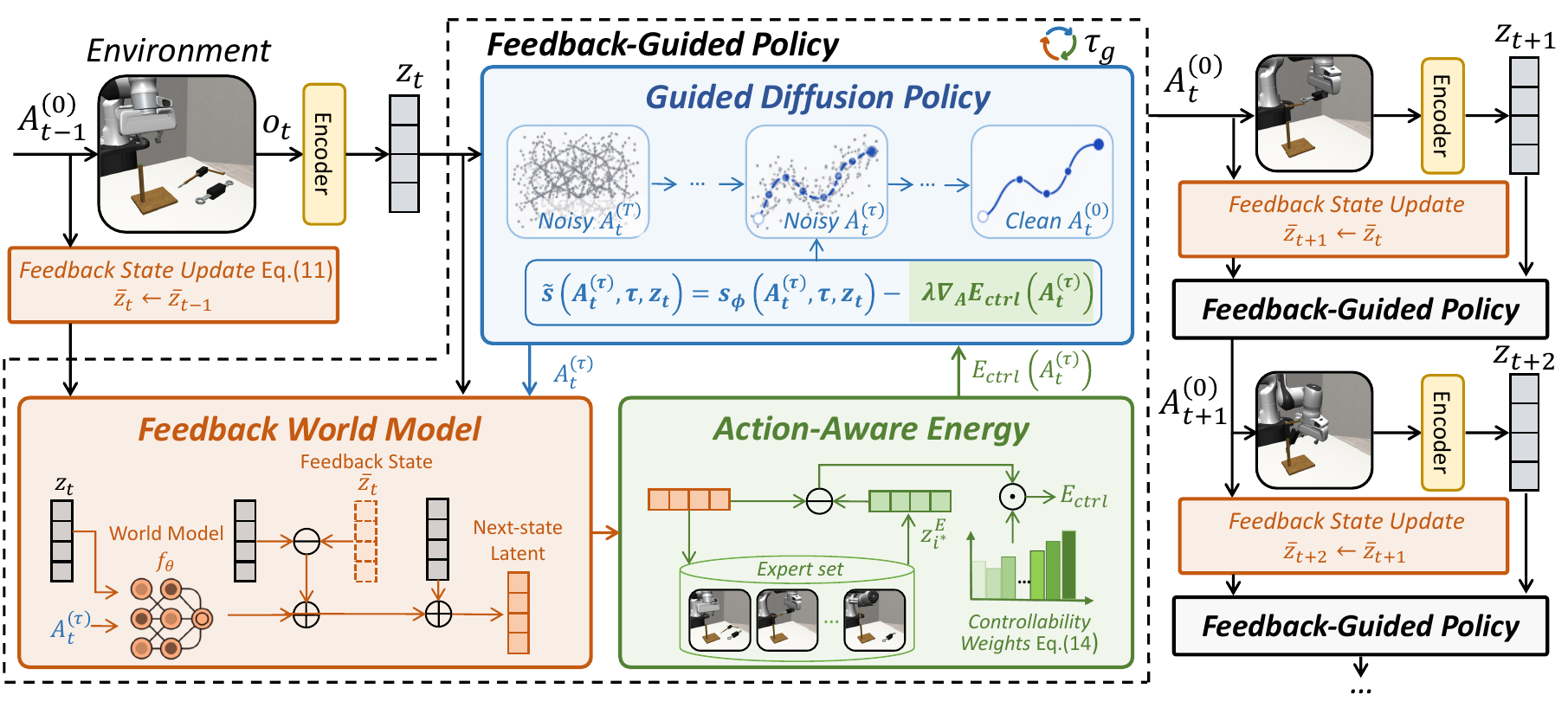}
    \caption{\textbf{Overview of the Feedback-Guided Policy.} During denoising,
the feedback world model predicts the future latent outcome of the current
action trajectory, and an action-aware energy guides the policy toward
expert-like, action-relevant states. After each execution, the new observation
updates the feedback state, which corrects subsequent feedback world model
predictions and forms an outer loop for reducing prediction drift during
inference.}
    \label{framwork overview}
\end{figure}

\section{Methodology}
\label{sec:methodology}
\subsection{Framework overview}
\label{sec:framework overview}

We first provide an overview of the proposed \emph{Feedback-Guided Policy}
(Fig.~\ref{framwork overview}), an inference framework that uses online
observation feedback to construct reliable and action-aware policy guidance. At each denoising step $\tau$, a candidate
action sequence $A_t^{(\tau)}$ is evaluated through a learned world model.
Following prior world model guidance methods~\citep{latent-policy-barrier},
the predicted next latent state $\hat z_{t+1}(A_t^{(\tau)})$ is matched to the
nearest expert latent in $\mathcal{Z}^E=\{z_i^E\}_{i=1}^N$ with $N$ samples:
\begin{equation}
    i^\star
    =
    \arg\min_i
    \left\|
    \hat z_{t+1}(A_t^{(\tau)}) - z_i^E
    \right\|_2^2 .
    \label{eq:nearest_expert}
\end{equation}
The corresponding guidance energy is then defined as the squared latent
distance to this nearest expert state:
\begin{equation}
    E(A_t^{(\tau)})
    =
    \left\|
    \hat z_{t+1}(A_t^{(\tau)}) - z_{i^\star}^E
    \right\|_2^2 .
    \label{eq:energy_Eq}
\end{equation}
This energy encourages candidate actions whose predicted outcomes are closer
to expert-like latent states. It is used to guide diffusion sampling as
\begin{equation}
    \tilde{s}
    =
    s_{\phi}
    -
    \lambda
    \nabla_{A_t^{(\tau)}}
    E(A_t^{(\tau)}),
    \label{eq:wm_guidance}
\end{equation}
where $\lambda$ controls the guidance strength, and 
$\nabla_{A_t^{(\tau)}}E(A_t^{(\tau)})$ denotes the gradient of the guidance
energy with respect to the candidate action sequence.

This standard formulation is limited by two factors: the predicted latent state
may drift under distribution shift, and uniform latent matching may include
action-irrelevant components in the guidance signal. We address these issues
with two complementary designs. First, a \emph{feedback world model} uses
online observations to correct latent predictions during inference. Second, an
\emph{action-aware energy} reweights latent dimensions according to their
estimated action controllability. Together, they produce guidance based on
predictions that are both dynamically reliable and action-relevant.
Algorithm~\ref{alg:fwm-inference} in Appendix~\ref{apendix: inference pipeline}
summarizes the full inference pipeline. In the following subsections, we detail
the feedback world model and action-aware energy, respectively.

\subsection{Feedback world model}
\label{sec:feedback_world_model}
As discussed in Sec.~\ref{sec:framework overview}, world-model-based guidance relies on the predicted latent $\hat{z}_{t+1}$ to construct the energy signal. 
In standard usage, the pretrained world model $f_{\theta}$ predicts in an open-loop manner according to Eq.~\eqref{world_model}; prediction errors under distribution shift can therefore bias guidance.

To improve prediction reliability, we introduce a \emph{feedback world model}
that uses real observations to correct latent predictions \emph{without
updating} the pretrained world model $f_\theta$. Following~\citep{jia2024feedback}, we reinterpret the
learned transition model in velocity form:
\begin{equation}
\label{eq:velocity-form}
    v_t(A_t^{(\tau)})
    \;\triangleq\;
    \frac{f_\theta(z_t, A_t^{(\tau)}) - z_t}{\delta t}=\frac{\hat{z}_{t+1}-z_t}{\delta t},
\end{equation}
where $\delta t$ is the environment update interval. In addition to the observed
latent state $z_t$, we maintain an auxiliary feedback state $\bar{z}_t$ (initialized as $z_0$) that
tracks the model's internal \textit{belief} about where the system should be after past
executed actions. The discrepancy
\begin{equation}
    e_t = z_t - \bar{z}_t
\end{equation}
therefore measures the accumulated prediction discrepancy up to the current
environment step.

During diffusion guidance at step $t$, this discrepancy is held fixed and used
to correct the score function for every noisy action $A_t^{(\tau)}$:
\begin{equation}
\label{eq:core-update-formula}
\left\{
\begin{aligned}
& \hat{v}_t(A_t^{(\tau)}) = v_t(A_t^{(\tau)}) + L e_t, \\
& z^{\mathrm{fb}}_{t+1}(A_t^{(\tau)})
    = z_t + \delta t \cdot \hat{v}_t(A_t^{(\tau)}),
\end{aligned}
\right.
\end{equation}
where $L$ is the positive-definite feedback gain. The corrected prediction
$z^{\mathrm{fb}}_{t+1}(A_t^{(\tau)})$ replaces the open-loop prediction $\hat{z}_{t+1}(A_t^{(\tau)})$ in the guidance
energy of Eq.~\eqref{eq:energy_Eq}. After the guided clean action $A_t^{(0)}$ is executed in the environment, the feedback state is propagated using the same action-conditioned corrected
velocity:
\begin{equation}
\label{eq:zbar-update}
    \bar{z}_{t+1}
    =
    \bar{z}_t + \delta t \cdot \hat{v}_t(A_t^{(0)}).
\end{equation}
After the environment returns the next observation $o_{t+1}$, we obtain the
corresponding observed latent state $z_{t+1}$. This latent state reflects the
actual transition reached after executing $A_t^{(0)}$, and is used to form the
next feedback residual:
\begin{equation}
\label{eq:residual error}
    e_{t+1}=z_{t+1}-\bar{z}_{t+1},
\end{equation}
which is carried forward and used to correct predictions at step $t{+}1$. Thus,
the action $A_t^{(0)}$ and the observation $z_{t+1}$ are aligned across one
environment transition: $A_t^{(0)}$ advances the internal \textit{belief} to
$\bar{z}_{t+1}$, and the subsequently observed $z_{t+1}$ calibrates the residual signal used at the next decision step.

\paragraph{Convergence guarantee.}
Assume a symmetric positive-definite gain $L$, a bounded latent-velocity
residual $\|\Delta_t\|\leq\gamma$ with $\Delta_t\to\Delta_\infty$, and,
for the discrete update, $\rho(I-\delta t L)<1$. Then
\begin{equation}
    \lim_{t\to\infty}\|z_t-\bar z_t\|
    \leq \frac{\gamma}{\lambda_{\min}(L)}.
\end{equation}
For $L=lI$, the bound is $\gamma/l$. The proof and
the corrected one-step prediction result appear in
Appendix~\ref{appendix: proof of convergence guarantee}.

\paragraph{Gain selection.}
We use $L=lI$ and compare a coarse set of gains using held-out prediction
MSE. For an update interval $\delta t$, the non-oscillatory homogeneous
range is $0<l\leq1/\delta t$; the effective gain is $\delta t l$.
Appendix~\ref{app:sensitivity} reports task settings and sensitivity of prediction
accuracy to $l$ (Table~\ref{tab:gain_sweep}).

\subsection{Action-aware guidance}
World-model-based guidance in Eq.~\eqref{eq:wm_guidance} depends not only on prediction accuracy, but also on whether the guidance objective emphasizes action-relevant state changes. 
Uniform latent matching is suboptimal because latent representations often entangle controllable factors, such as robot-object geometry, with weakly controllable factors, such as background appearance and lighting. 
We therefore propose an \emph{action-aware energy} that reweights latent dimensions according to their controllability.

Specifically, we estimate the controllability of the $j$-th latent dimension by its
\emph{counterfactual variance} under the learned world model:
\begin{equation}
\label{eq:weight_calcu}
    w_j
    =
    \frac{1}{M}
    \sum_{i=1}^{M}
    \mathrm{Var}_{a \sim \mathcal{A}_{\mathrm{demo}}}
    \!\left(
    [f_{\theta}(z_i,\,a)]_j
    \right),
    \quad j \in \{1,\dots,D\},
\end{equation}
where $\{z_i\}_{i=1}^{M}$ are observations drawn at uniformly spaced
indices along the demonstration trajectory, and the variance is taken
over actions sampled from the demonstration action pool
$\mathcal{A}_{\mathrm{demo}}$.
A larger $w_j$ indicates that the corresponding latent dimension is
more responsive to action perturbations and is therefore more relevant
for action guidance.
In practice we approximate~\eqref{eq:weight_calcu} with $M{=}200$
states and 32 action samples per state.

We normalize the weights and apply a soft interpolation to preserve
the overall scale of the energy:
\begin{equation}
\label{eq:weight_normalize}
    \bar w_j
    =
    \frac{w_j}{\frac{1}{D}\sum_{k=1}^{D} w_k + \epsilon},
    \qquad
    w_j^{(\beta)}
    =
    1 + \beta\,(\bar w_j - 1),
\end{equation}
where $\epsilon = 10^{-8}$ is a small constant added for numerical
stability that prevents division by zero in the unlikely event that
the average counterfactual variance vanishes, and $\beta \in [0,1]$
controls the strength of controllability modulation.
The final guidance loss replaces the standard MSE with the
controllability-weighted form
\begin{equation}
\label{eq:Ectrl_calcu}
    E_{\mathrm{ctrl}}(A^{(\tau)}_t)
    =
    \sum_{j=1}^{D} w_j^{(\beta)} \,(z^{\mathrm{fb}}_{t+1,j}(A_t^{(\tau)}) - z_{i^\star,j}^E)^2.
\end{equation}
The gradient $\nabla_A E_{\mathrm{ctrl}}$ is then used as the guidance term in Eq.~\eqref{eq:wm_guidance}, so action-sensitive latent dimensions contribute more strongly to denoising guidance, while action-invariant dimensions are down-weighted.

\section{Experiments}
\label{sec:experiments}
We conduct experiments mainly to answer two questions. First, can the proposed feedback world model reduce prediction error online in OOD settings? Second, when combined with controllability-aware guidance, can the improved feedback world model further enhance the OOD success rate of diffusion policies? In simulation, we evaluate robustness to perturbations in robot initial states, lighting, background textures, object layouts, and sensor noise. On real robots, we test pick-and-place and drawer-opening tasks under initial-pose shifts. We also apply feedback to the frozen pretrained foundation model Fast-WAM-Joint across the full LIBERO-10 suite in LIBERO-Plus to assess its plug-and-play applicability.

\begin{figure}
    \centering
    \includegraphics[width=1\linewidth]{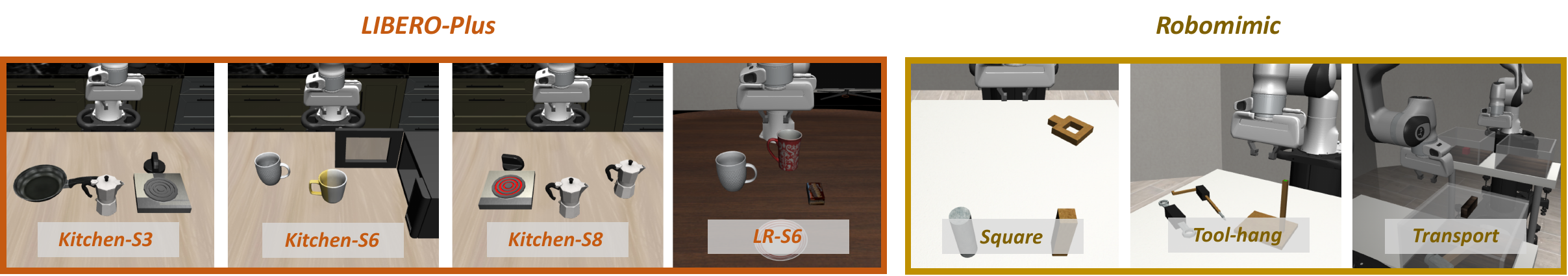}
    \caption{\textbf{Simulated tasks.} For LIBERO-Plus, we evaluate four representative single-task settings from the \textit{Kitchen} and \textit{Living Room} (LR) scenes, with further details provided in Appendix~\ref{appendix: libero_plus_tasks}. For Robomimic tasks, we evaluate three representative tasks including Transport, Square and Tool-Hang.}
    \label{fig:simulated tasks}
\end{figure}

\begin{figure}[t]
\centering
\includegraphics[width=\linewidth]{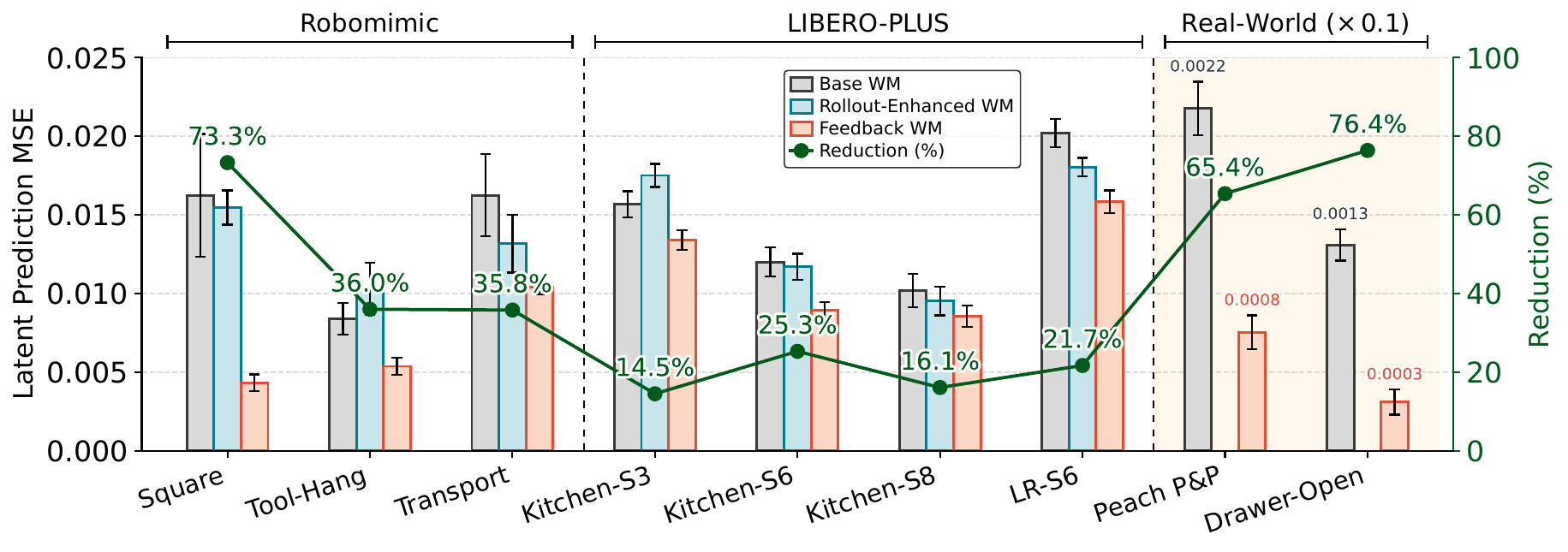}
\caption{\textbf{Latent prediction MSE under robot-initial-state OOD.}
Simulation: step-pooled means $\pm$ episode-cluster bootstrap SE;
real world: episode-equal means $\pm$ SEM. The green curve shows the relative
reduction from Base to Feedback. Real-world bars and error bars are magnified
$10\times$ for display.
Counts and estimators are detailed
in Appendix~\ref{app:statistics}.}
\label{fig:wm_mse}
\end{figure}

\begin{figure}[t]
    \centering
    \includegraphics[width=\linewidth]{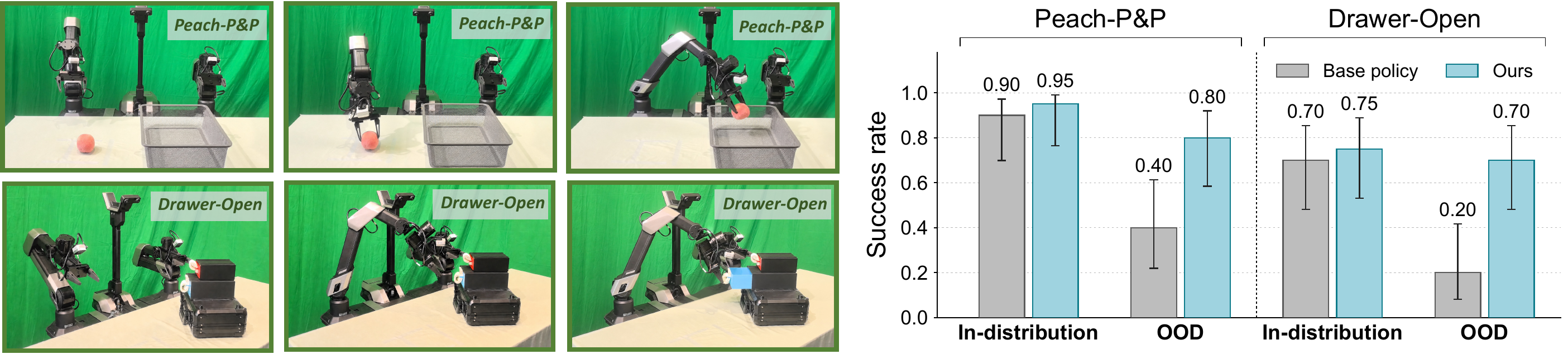}
    \caption{\textbf{Real-world tasks and results.} We deploy the policy on Peach-P\&P and Drawer-Open tasks under both in-distribution and out-of-distribution initial states. Each point uses 20 binary trials; error bars are 95\% Wilson
    intervals.}
    \label{fig: real-world tasks and results}
\end{figure}

\begin{table*}[t]
\centering
\small
\setlength{\tabcolsep}{4.2pt}
\caption{
\textbf{Success rates on simulated OOD manipulation tasks.}
Baselines include \textit{Base} policy, \textit{Mixed BC} using expert and rollout data, 
\textit{Filtered BC} removing failed rollouts, \textit{Std. Guid.} using the base world model, 
and \textit{LPB-style Rollout Guidance} using a rollout-enhanced world model. 
Best and second-best results are bolded and underlined. Rollout details appear in Appendix~\ref{appendix: rollout data collection}.
}
\label{tab:sim_success}
\vspace{0.5em}
\begin{tabular}{llcccccc}
\toprule
Benchmark & Task 
& Base 
& Mixed BC 
& Filtered BC 
& Std. Guid. 
& Rollout Guid. 
& Ours \\
\midrule
\multirow{3}{*}{Robomimic}
& Square    
& 0.36 & 0.08 & 0.12 & 0.40 & \underline{0.44} & \textbf{0.46} \\
& Tool-Hang 
& 0.27 & 0.22 & 0.22 & 0.22 & \underline{0.31} & \textbf{0.36} \\
& Transport 
& 0.34 & 0.26 & 0.36 & 0.46 & \underline{0.52} & \textbf{0.60} \\
\midrule
\multirow{4}{*}{LIBERO-Plus}
& Kitchen-S3 
& 0.55 & 0.57 & \textbf{0.70} & 0.61 & 0.50 & \underline{0.68} \\
& Kitchen-S6 
& 0.45 & 0.48 & \underline{0.54} & 0.48 & 0.50 & \textbf{0.55} \\
& Kitchen-S8 
& 0.59 & 0.42 & 0.35 & \underline{0.62} & \underline{0.62} & \textbf{0.65} \\
& LR-S6      
& \underline{0.26} & 0.24 & 0.24 & 0.21 & 0.24 & \textbf{0.36} \\
\midrule
\multicolumn{2}{l}{Average} 
& 0.40 & 0.32 & 0.36 & 0.43 & \underline{0.45} & \textbf{0.52} \\
\bottomrule
\end{tabular}
\vspace{-0.5em}
\end{table*}

\subsection{Experimental setup}

\paragraph{Simulation benchmarks.}
We evaluate our method on seven simulated manipulation tasks from Robomimic~\citep{robomimic} and LIBERO-Plus~\citep{libero-plus}, as shown in Fig.~\ref{fig:simulated tasks}. 
For Robomimic, we use three representative tasks, Square, Tool-Hang, and Transport, and train the diffusion policy in a low-data regime using 20\% of the official demonstrations. 
OOD evaluation on Robomimic perturbs the robot's initial state. For LIBERO-Plus, we use four single-task settings from LIBERO-10 and evaluate all five perturbation categories listed above.
For each simulated task, we train a base world model on the corresponding expert demonstrations and keep its parameters fixed during evaluation.
Task selection and perturbation protocols are detailed in Appendices~\ref{appendix: Robomimic OOD setting}, \ref{appendix: libero_plus_tasks}, and~\ref{app:broader_ood}; statistical reporting and the pretrained-model setup appear in Appendices~\ref{app:statistics} and~\ref{app:foundation}, respectively.

\paragraph{Real-world tasks.}
We further evaluate on two real-world manipulation tasks, Peach-P\&P and Drawer-Open, as shown in Fig.~\ref{fig: real-world tasks and results}. 
Peach-P\&P requires picking a peach from the table and placing it into a target box, while Drawer-Open requires handle localization and contact-rich pulling with a deformable rubber handle. 
We collect 50 and 65 expert trajectories for the two tasks, respectively, and train both the diffusion policy and world model from scratch. 
At test time, we reset the robot arm to abnormal initial poses outside the training distribution, yielding a real-world robot-initial-state OOD setting.

\subsection{Experimental results}
\label{sec:experimental_results}

\subsubsection{Feedback world model improves online prediction.}
We first evaluate whether the proposed feedback mechanism improves world model prediction under robot initial-state OOD shifts. 
Fig.~\ref{fig:wm_mse} shows latent prediction MSE under simulated and real-world OOD settings. 
The base world model incurs large errors under perturbed initial states.
By correcting its internal predictive state with online observations, our feedback world model reduces mean prediction error without extra training data or parameter updates. 
The reduction reaches $73.3\%$ on Robomimic Square and $65.4\%$/$76.4\%$ on the real-world Peach-P\&P and Drawer-Open tasks, confirming improved prediction reliability under OOD conditions.

Controlled action interventions and physical-state probes assess the
action dependence and physical relevance of these predictions
(Appendix~\ref{app:validation}). Comparisons with Kalman and learned
filtering baselines appear in Appendix~\ref{app:filtering};
Appendix~\ref{app:residual} analyzes how the model residual varies
during execution.

\subsubsection{Action-aware guidance improves controllable policy guidance.}
\begin{figure}[t]
    \centering    \includegraphics[width=1\linewidth]{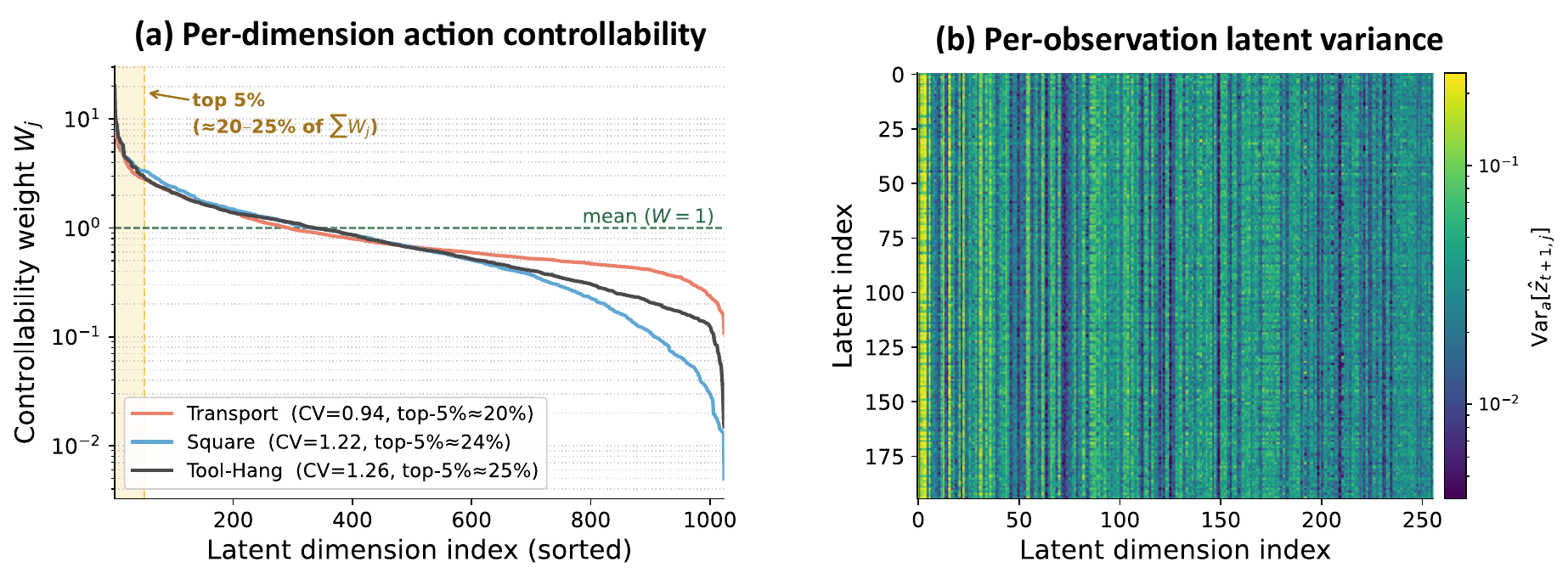}
    \caption{\textbf{Action controllability of the world model latent.}
\textbf{(a)} Sorted controllability weights $w_j$ on three Robomimic tasks, shown in log scale.
\textbf{(b)} Counterfactual action variance $\mathrm{Var}_a[\hat z_{t+1,j}]$ on Transport, computed over sampled observations and 256 selected latent dimensions. 
Persistent vertical bands indicate dimensions that remain action-responsive across observations, supporting the use of an offline-frozen $W$ at inference time.
}
    \label{fig: action-aware visualize}
\end{figure}

While the feedback world model improves prediction reliability, effective policy guidance also requires focusing the guidance objective on action-controllable state components. 
To examine this, we visualize the distribution of action controllability in the world model latent space (Fig.~\ref{fig: action-aware visualize}). 
Across three Robomimic tasks, the sorted weights show a clear long-tailed pattern: the top $5\%$ of dimensions contribute about $20$--$25\%$ of the total controllability weight, while many dimensions have much smaller weights. 
The per-latent-observation variance map further shows that high-weight dimensions remain consistently action-responsive across observations. 
These results reveal strong action anisotropy in the latent representation, supporting our design choice to emphasize action-controllable dimensions during guidance.

We further ablate action-aware guidance on Robomimic tasks, as shown in Table~\ref{tab:controllability ablation}. 
Compared with guidance using only the feedback world model, adding action-aware weighting further improves policy success rates. 
By downweighting action-irrelevant visual components, action-aware guidance focuses action optimization on task-relevant state changes and improves downstream policy success.

\subsubsection{Combined guidance improves OOD performance across benchmarks.}
Having verified the effectiveness of feedback correction and action-aware weighting separately, we next evaluate their combined effect on policy performance across broader simulated and real-world OOD settings. 
Table~\ref{tab:sim_success} and Fig.~\ref{fig: real-world tasks and results} report policy success rates in simulated and real-world OOD settings. 
The simulated results average $52\%$ for our method, compared with $40\%$ for the base policy, $43\%$ for standard guidance and $45\%$ for LPB-style rollout guidance.
Although Filtered BC slightly outperforms ours on Kitchen-S3, rollout-based augmentation is less stable overall: Mixed BC often degrades performance, and rollout-enhanced guidance improves the average success rate only moderately compared with standard guidance. 
This suggests that simply adding rollout data does not necessarily resolve OOD generalization, since rollout trajectories may contain failure cases, biased state distributions, or insufficient coverage of the final test-time perturbations.

Real-world results further confirm the benefit under robot initial-state shifts. 
Under in-distribution states, our method slightly improves the already strong base policy, from $90\%$ to $95\%$ on Peach-P\&P and from $70\%$ to $75\%$ on Drawer-Open. 
Under OOD states, the gains become much larger, improving success from $40\%$ to $80\%$ and from $20\%$ to $70\%$, respectively. 
Overall, the results suggest that reliable next-state prediction from the feedback world model, together with fine-grained action-aware guidance, leads to more robust policy performance under distribution shift.

\begin{table}[t]
\centering
\small
\setlength{\tabcolsep}{8pt}
\caption{
\textbf{Ablation study.}
\textit{+ Feedback WM} uses the feedback world model for guidance without per-dimension weighting;
\textit{+ Action-aware} additionally uses the offline counterfactual-variance score $W$.
}
\label{tab:controllability ablation}
\vspace{0.5em}
\begin{tabular}{lccc}
\toprule
Method & Square & Transport & Tool-Hang \\
\midrule
Base & 0.36 & 0.34  & 0.27 \\
+ Feedback WM & 0.40 & 0.54 & 0.32 \\
+ Action-aware & \textbf{0.46} & \textbf{0.60} & \textbf{0.36} \\
\bottomrule
\end{tabular}
\vspace{-0.5em}
\end{table}

\subsubsection{Robustness across LIBERO-Plus perturbations.}
We evaluate the four LIBERO-Plus tasks under perturbations to robot
initial states, lighting, background textures, object layout, and sensor
noise. Across these five categories, pooled success increases by
$4.5$--$9.9$ percentage points and latent MSE decreases by
$6.8$--$47.8\%$ (Table~\ref{tab:broader_ood}). Our method matches or
exceeds the base policy in 18 of the 20 task--perturbation settings.
Appendix~\ref{app:broader_ood} reports the results for each setting.

\begin{table}[t]
\centering\small
\caption{\textbf{LIBERO-Plus evaluation across five perturbation categories.}
Success rates are pooled within perturbation categories; MSE reduction is
relative to the base world model. $\Delta$SR is in percentage points (pp).
}
\label{tab:broader_ood}
\begin{tabular}{lrrrr}
\toprule
Perturbation & Base SR & Ours SR & $\Delta$SR & MSE reduction\\
\midrule
Robot initial states & 45.7\% & 55.6\% & +9.9 & 19.4\%\\
Lighting & 42.2\% & 46.7\% & +4.5 & 47.8\%\\
Background textures & 59.5\% & 64.9\% & +5.4 & 42.2\%\\
Object layout & 67.2\% & 75.2\% & +8.0 & 12.3\%\\
Sensor noise & 81.8\% & 86.6\% & +4.8 & 6.8\%\\
\bottomrule
\end{tabular}
\end{table}

\subsubsection{Transfer to a pretrained world-action model.}
To test the plug-and-play applicability of feedback, we apply our feedback
mechanism to Fast-WAM-Joint~\citep{fast-wam}, built on the pretrained
Wan2.2-5B backbone. The model uses two-camera observations and a single
multi-task policy across the full LIBERO-10 suite. We keep its architecture
and learned parameters frozen, introducing feedback at inference time
without retraining or additional training data.

\begin{figure}[htbp]
\centering
\includegraphics[width=\linewidth]{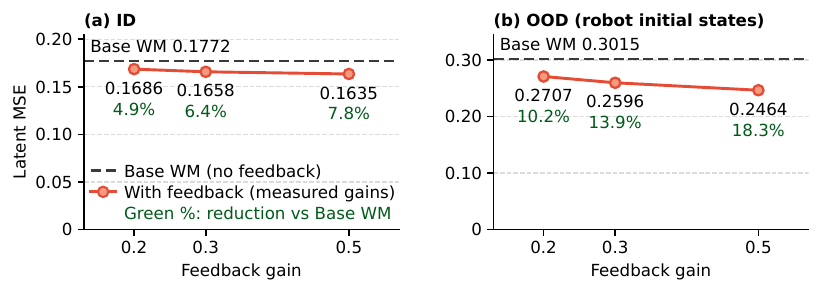}
\caption{\textbf{Feedback improves frozen Fast-WAM-Joint predictions.}
Latent MSE on LIBERO-10 under ID and robot-initial-state OOD conditions.
Dashed lines show the base model; markers show feedback at each tested
gain. Green labels give the relative MSE reduction.}
\label{fig:foundation_gain}
\end{figure}

\paragraph{World model prediction.}
We evaluate 7,553 paired predictions over all 393 robot-initial-state
OOD variants and 1,203 ID predictions from 100 episodes.
Fig.~\ref{fig:foundation_gain} shows that feedback reduces latent MSE
under both ID and OOD conditions at all three tested gains.
At gain $0.5$, ID MSE decreases from $0.1772$ to $0.1635$ ($7.8\%$),
and OOD MSE decreases from $0.3015$ to $0.2464$ ($18.3\%$).
Against VAE-reconstructed observations, OOD pixel MSE decreases by
$18.0\%$ and PSNR increases by $0.86$ dB. Together, these results show that inference-time feedback
improves the predictions of a pretrained foundation model in both latent
and image space, while preserving its learned parameters.

\paragraph{Policy performance.}
On the LIBERO-Plus Robot Initial States OOD evaluation across LIBERO-10,
our method improves the success rate of Fast-WAM-Joint from $77.1\%$ to $81.4\%$, a gain of $4.3$
percentage points. The prediction and policy results demonstrate that
our method can be applied to an existing pretrained world-action model
without updating its parameters. Appendix~\ref{app:foundation} provides
the evaluation details.

\section{Conclusion}
\label{sec:conclusion}
In this work, we introduce a feedback world model for robust test-time guidance of diffusion policies. 
Instead of using a pretrained world model as a static open-loop predictor, our method exploits real observations collected during policy execution to correct the model's latent predictive state online, improving prediction reliability without additional training data or parameter updates. 
We further propose action-aware guidance, which weights latent prediction errors by offline-estimated controllability to focus guidance on action-sensitive state components. 
Extensive simulated and real-world experiments show that combining feedback correction with action-aware guidance leads to more robust policy execution in OOD manipulation settings. 
\paragraph{Limitations and future work.}
Similar to prior inference-time guidance methods for diffusion policies~\citep{latent-policy-barrier, progressVLA}, our approach has higher inference latency than an unguided base diffusion policy because guidance is applied during denoising. 
Since the feedback update is analytic and the controllability weights are computed offline, most overhead comes from the guided-diffusion paradigm rather than from our proposed components. 
Future work can apply these ideas to more efficient guidance frameworks and latency-sensitive robotic tasks.

\clearpage
\begin{ack}
This work is supported by MOE Singapore Tier 1 Grant RG83/25, RS36/24, and a Start-up Grant from Nanyang Technological University. The authors declare no competing interests.
\end{ack}

\bibliographystyle{plainnat} 
\bibliography{references} 

\clearpage
\appendix

\section{Proof of the convergence guarantee}
\label{appendix: proof of convergence guarantee}
We provide the convergence analysis of the proposed feedback world model.
Let $z_t=\psi(O_t)$ denote the observed latent state and let $\bar z_t$
denote the auxiliary feedback state. For the executed action $A_t^{(0)}$, we
write the true latent transition in velocity form as
\begin{equation}
\label{eq:true-latent-velocity-proof}
    z_{t+1}
    =
    z_t+\delta t\, v_t^\star(A_t^{(0)}),
\end{equation}
where $v_t^\star(A_t^{(0)})$ is the true latent velocity induced by the
environment. The learned world model gives
\begin{equation}
\label{eq:learned-latent-velocity-proof}
    v_t(A_t^{(0)})
    =
    \frac{f_\theta(z_t,A_t^{(0)})-z_t}{\delta t}.
\end{equation}
We assume that the learned latent velocity has a bounded and asymptotically
convergent residual error:
\begin{equation}
\label{eq:bounded-residual-proof}
    v_t^\star(A_t^{(0)})
    =
    v_t(A_t^{(0)})+\Delta_t,
    \qquad
    \|\Delta_t\|\leq \gamma,
    \qquad
    \Delta_t\to \Delta_\infty ,
\end{equation}
where $\gamma>0$ is an unknown constant and
$\|\Delta_\infty\|\leq\gamma$. This assumption states that the learned latent
dynamics may have a nonzero residual error, but the residual remains bounded
and approaches a steady limiting disturbance along the deployment trajectory.

The feedback world model corrects the learned velocity by
\begin{equation}
\label{eq:corrected-velocity-proof}
    \hat v_t(A_t^{(0)})
    =
    v_t(A_t^{(0)})+L(z_t-\bar z_t),
\end{equation}
where $L=L^\top\succ 0$ is the feedback gain. The auxiliary feedback state is updated as
\begin{equation}
\label{eq:feedback-state-update-proof}
    \bar z_{t+1}
    =
    \bar z_t+\delta t\,\hat v_t(A_t^{(0)}).
\end{equation}

Define the feedback estimation error as
\begin{equation}
\label{eq:error-definition-proof}
    e_t=z_t-\bar z_t .
\end{equation}
Combining Eq.~\eqref{eq:true-latent-velocity-proof},
Eq.~\eqref{eq:bounded-residual-proof},
Eq.~\eqref{eq:corrected-velocity-proof}, and
Eq.~\eqref{eq:feedback-state-update-proof}, we obtain
\begin{align}
    e_{t+1}
    &=
    z_{t+1}-\bar z_{t+1}
    \nonumber \\
    &=
    z_t+\delta t\big(v_t(A_t^{(0)})+\Delta_t\big)
    -
    \left[
    \bar z_t+\delta t
    \big(v_t(A_t^{(0)})+L(z_t-\bar z_t)\big)
    \right]
    \nonumber \\
    &=
    (I-\delta t L)e_t+\delta t\,\Delta_t .
\label{eq:discrete-error-dynamics-proof}
\end{align}
Thus, the feedback residual follows a stable linear system driven by the
latent dynamics residual.

We first analyze the continuous-time counterpart of
Eq.~\eqref{eq:discrete-error-dynamics-proof}, which is
\begin{equation}
\label{eq:continuous-error-dynamics-proof}
    \dot e(t)
    =
    -Le(t)+\Delta(t),
    \qquad
    \|\Delta(t)\|\leq \gamma,
    \qquad
    \Delta(t)\to \Delta_\infty .
\end{equation}
Since $L\succ 0$, the homogeneous system $\dot e(t)=-Le(t)$ is exponentially
stable. The solution of Eq.~\eqref{eq:continuous-error-dynamics-proof} is
\begin{equation}
\label{eq:continuous-solution-proof}
    e(t)
    =
    e^{-Lt}e(0)
    +
    \int_0^t e^{-L(t-s)}\Delta(s)\,ds .
\end{equation}
The first term converges to zero because all eigenvalues of $L$ are positive:
\begin{equation}
    \lim_{t\to\infty} e^{-Lt}e(0)=0 .
\end{equation}
For the second term, since $\Delta(t)\to\Delta_\infty$, the forced response of
the stable linear system converges to the equilibrium induced by the limiting
residual:
\begin{equation}
\label{eq:continuous-equilibrium-proof}
    \lim_{t\to\infty} e(t)
    =
    L^{-1}\Delta_\infty .
\end{equation}
Therefore,
\begin{equation}
\label{eq:observer-bound-proof}
    \lim_{t\to\infty}
    \|e(t)\|
    =
    \|L^{-1}\Delta_\infty\|
    \leq
    \|L^{-1}\|_2\|\Delta_\infty\|
    \leq
    \frac{\gamma}{\lambda_{\min}(L)} .
\end{equation}
Since $e(t)=z(t)-\bar z(t)$, we obtain
\begin{equation}
\label{eq:continuous-final-bound-proof}
    \lim_{t\to\infty}
    \|z(t)-\bar z(t)\|
    \leq
    \frac{\gamma}{\lambda_{\min}(L)} .
\end{equation}
For the scalar-gain case $L=lI$ with $l>0$, this reduces to
\begin{equation}
\label{eq:scalar-observer-bound-proof}
    \lim_{t\to\infty}
    \|z(t)-\bar z(t)\|
    \leq
    \frac{\gamma}{l}.
\end{equation}

For the discrete-time update used in our implementation, Eq.~\eqref{eq:discrete-error-dynamics-proof}
can be written as
\begin{equation}
\label{eq:discrete-linear-system-proof}
    e_{t+1}=Ae_t+\delta t\,\Delta_t,
    \qquad
    A=I-\delta t L .
\end{equation}
If the feedback gain satisfies
\begin{equation}
\label{eq:discrete-stability-condition-proof}
    \rho(A)=\rho(I-\delta t L)<1,
\end{equation}
where $\rho(\cdot)$ denotes the spectral radius, then the discrete error system
is exponentially stable in the absence of residual input. Unrolling
Eq.~\eqref{eq:discrete-linear-system-proof} gives
\begin{equation}
\label{eq:discrete-unroll-proof}
    e_t
    =
    A^t e_0
    +
    \delta t
    \sum_{k=0}^{t-1}
    A^{t-1-k}\Delta_k .
\end{equation}
Because $\rho(A)<1$, we have $A^t e_0\to 0$. Moreover, since
$\Delta_t\to\Delta_\infty$, the input response converges to the steady-state
response of the stable linear system:
\begin{equation}
\label{eq:discrete-error-limit-proof}
    \lim_{t\to\infty} e_t
    =
    \delta t
    \sum_{j=0}^{\infty}
    A^j\Delta_\infty
    =
    \delta t (I-A)^{-1}\Delta_\infty .
\end{equation}
Since $I-A=\delta t L$, this becomes
\begin{equation}
\label{eq:discrete-equilibrium-proof}
    \lim_{t\to\infty} e_t
    =
    L^{-1}\Delta_\infty .
\end{equation}
Therefore,
\begin{equation}
\label{eq:discrete-matrix-bound-proof}
    \lim_{t\to\infty}
    \|e_t\|
    =
    \|L^{-1}\Delta_\infty\|
    \leq
    \frac{\gamma}{\lambda_{\min}(L)} .
\end{equation}
Equivalently,
\begin{equation}
\label{eq:discrete-final-bound-proof}
    \lim_{t\to\infty}
    \|z_t-\bar z_t\|
    \leq
    \frac{\gamma}{\lambda_{\min}(L)} .
\end{equation}

In particular, when $L=lI$ and $0<\delta t\,l<2$, the stability condition
$\rho(I-\delta t L)<1$ holds. If additionally $0<\delta t\,l\leq 1$, then
\begin{equation}
    \|e_{t+1}\|
    \leq
    (1-\delta t\,l)\|e_t\|+\delta t\,\|\Delta_t\|.
\end{equation}
Since $\Delta_t\to\Delta_\infty$, the exact scalar-gain error dynamics
converges to
\begin{equation}
\label{eq:discrete-scalar-limit-proof}
    \lim_{t\to\infty} e_t
    =
    \frac{1}{l}\Delta_\infty ,
\end{equation}
and hence
\begin{equation}
\label{eq:discrete-scalar-bound-proof}
    \lim_{t\to\infty}
    \|z_t-\bar z_t\|
    =
    \frac{1}{l}\|\Delta_\infty\|
    \leq
    \frac{\gamma}{l}.
\end{equation}

Finally, we relate this observer convergence result to the corrected one-step
prediction used for policy guidance. For the executed action $A_t^{(0)}$, the
corrected prediction is
\begin{equation}
\label{eq:corrected-prediction-proof}
    z^{\mathrm{fb}}_{t+1}(A_t^{(0)})
    =
    z_t+\delta t
    \big(v_t(A_t^{(0)})+Le_t\big).
\end{equation}
Using Eq.~\eqref{eq:true-latent-velocity-proof} and
Eq.~\eqref{eq:bounded-residual-proof}, its prediction error satisfies
\begin{align}
    z^{\mathrm{fb}}_{t+1}(A_t^{(0)})-z_{t+1}
    &=
    z_t+\delta t\big(v_t(A_t^{(0)})+Le_t\big)
    -
    z_t-\delta t\big(v_t(A_t^{(0)})+\Delta_t\big)
    \nonumber \\
    &=
    \delta t\big(Le_t-\Delta_t\big).
\label{eq:corrected-prediction-error-proof}
\end{align}
Since $e_t\to L^{-1}\Delta_\infty$ and $\Delta_t\to\Delta_\infty$, we have
\begin{equation}
\label{eq:corrected-prediction-error-limit-proof}
    \lim_{t\to\infty}
    \left(
    Le_t-\Delta_t
    \right)
    =
    L L^{-1}\Delta_\infty-\Delta_\infty
    =
    0 .
\end{equation}
Therefore,
\begin{equation}
\label{eq:corrected-prediction-bound-proof}
    \lim_{t\to\infty}
    \left\|
    z^{\mathrm{fb}}_{t+1}(A_t^{(0)})-z_{t+1}
    \right\|
    =
    0 .
\end{equation}

This result shows that, under a bounded and asymptotically convergent latent
dynamics residual, the auxiliary feedback state converges to a bounded
neighborhood of the observed latent state, with radius controlled by the
residual level and the feedback gain. Moreover, the corrected one-step
prediction used for guidance becomes asymptotically unbiased for the executed
action. Hence, the feedback world model prevents deployment-time prediction
bias from growing unboundedly, and the remaining observer bias is controlled by
the learned latent dynamics residual and the feedback gain.

\section{Robomimic OOD perturbation settings}
\label{appendix: Robomimic OOD setting}

Standard Robomimic evaluation initializes test episodes from the same
distribution as the training demonstrations. As a result, both the policy and
the world model are usually evaluated on familiar robot configurations. To test
robustness under a realistic deployment shift, we construct an OOD protocol that
perturbs the robot's initial joint configuration. 

For each test episode, we start from the initial simulator state of  Robomimic
demonstrations and perturb seven arm joint angles:
\begin{equation}
\tilde{\mathbf q}^{(i)}
=
\mathrm{clip}\left(
\mathbf q^{(i)}+\boldsymbol{\epsilon}^{(i)},
\mathbf q_{\min},
\mathbf q_{\max}
\right),
\qquad
\boldsymbol{\epsilon}^{(i)}
\sim
\mathcal N(\mathbf 0,\sigma^2 I_7),
\label{eq:robomimic-ood-perturb}
\end{equation}
where $\mathbf q^{(i)}$ is the demonstration initial joint configuration and
$[\mathbf q_{\min},\mathbf q_{\max}]$ are the Panda joint limits. For the
dual-arm Transport task, the same perturbation procedure is applied
independently to each arm. All other simulator state variables are left
unchanged.

\begin{figure}[t]
    \centering
    \includegraphics[width=\linewidth]{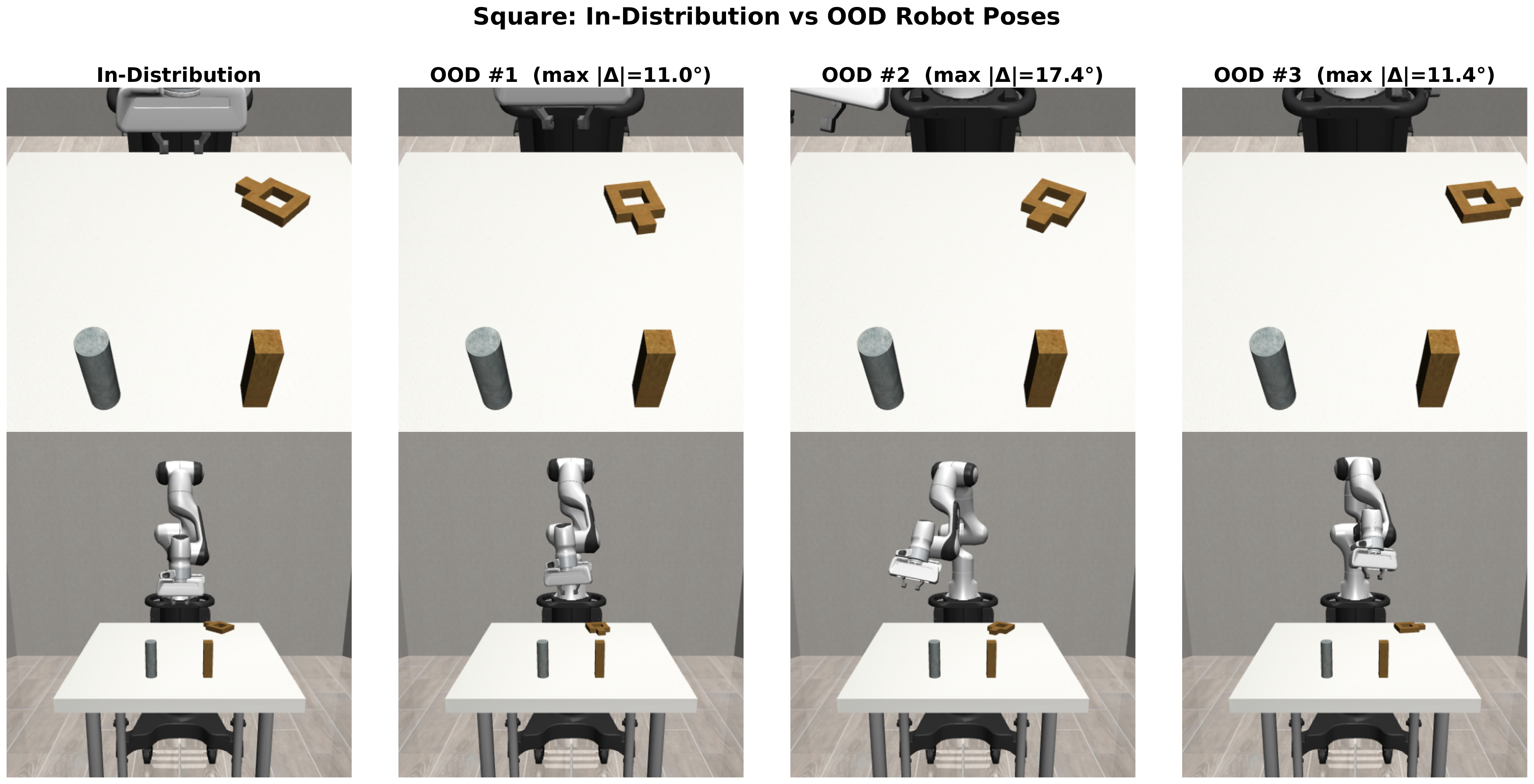}
    \caption{\textbf{Square OOD initial states.} The leftmost column shows the
    in-distribution demonstration start. The remaining columns show OOD starts
    generated by perturbing only the robot arm joints. Top row: agentview camera;
    bottom row: front-view camera.}
    \label{fig:robomimic-ood-square}
\end{figure}

\begin{figure}[t]
    \centering
    \includegraphics[width=\linewidth]{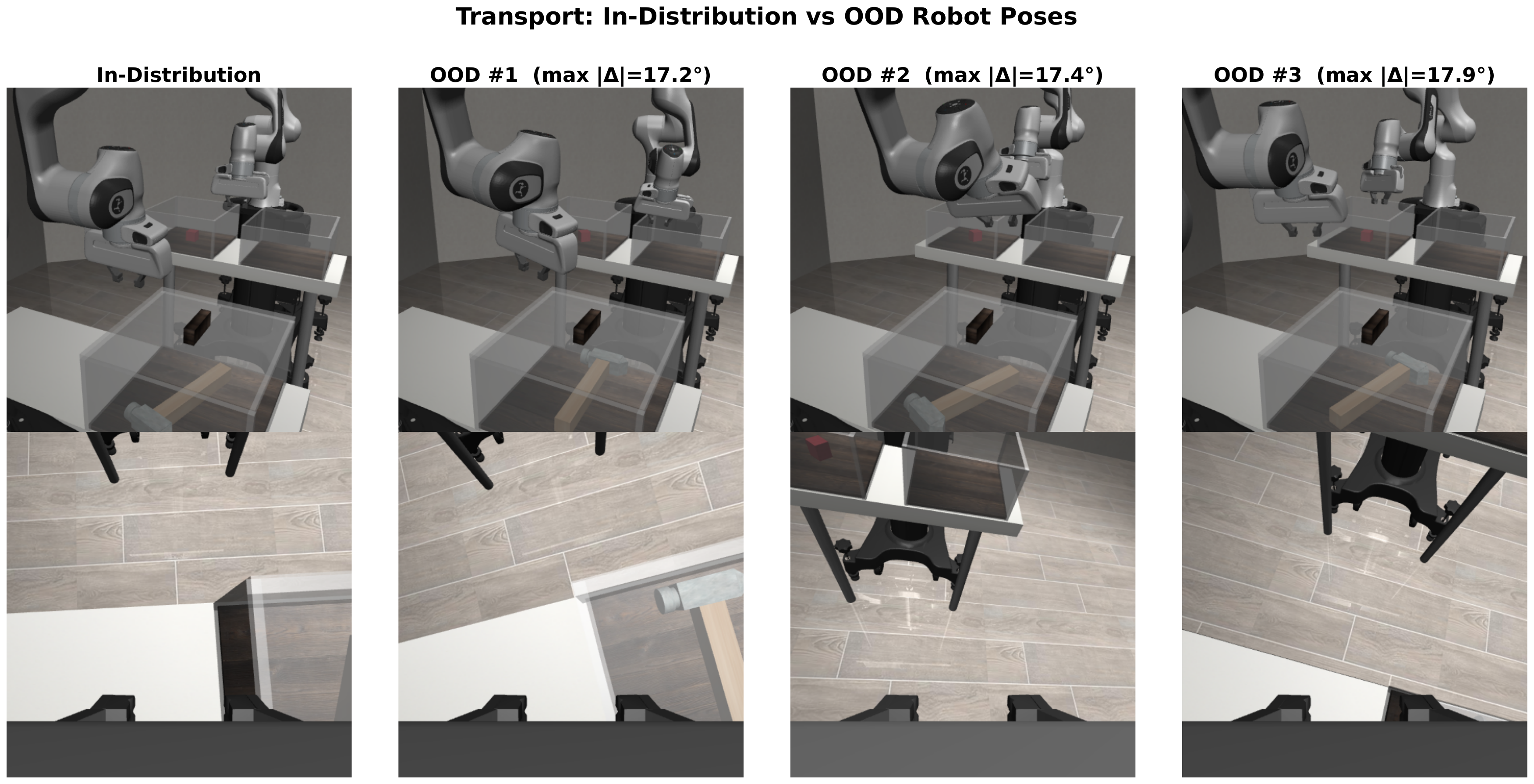}
    \caption{\textbf{Transport OOD initial states.} The layout follows
    Fig.~\ref{fig:robomimic-ood-square}. For this dual-arm task, joint
    perturbations are applied independently to the two arms. Top row:
    \texttt{shouldercamera0}; bottom row: robot-0 wrist camera.}
    \label{fig:robomimic-ood-transport}
\end{figure}

\begin{figure}[h]
    \centering
    \includegraphics[width=\linewidth]{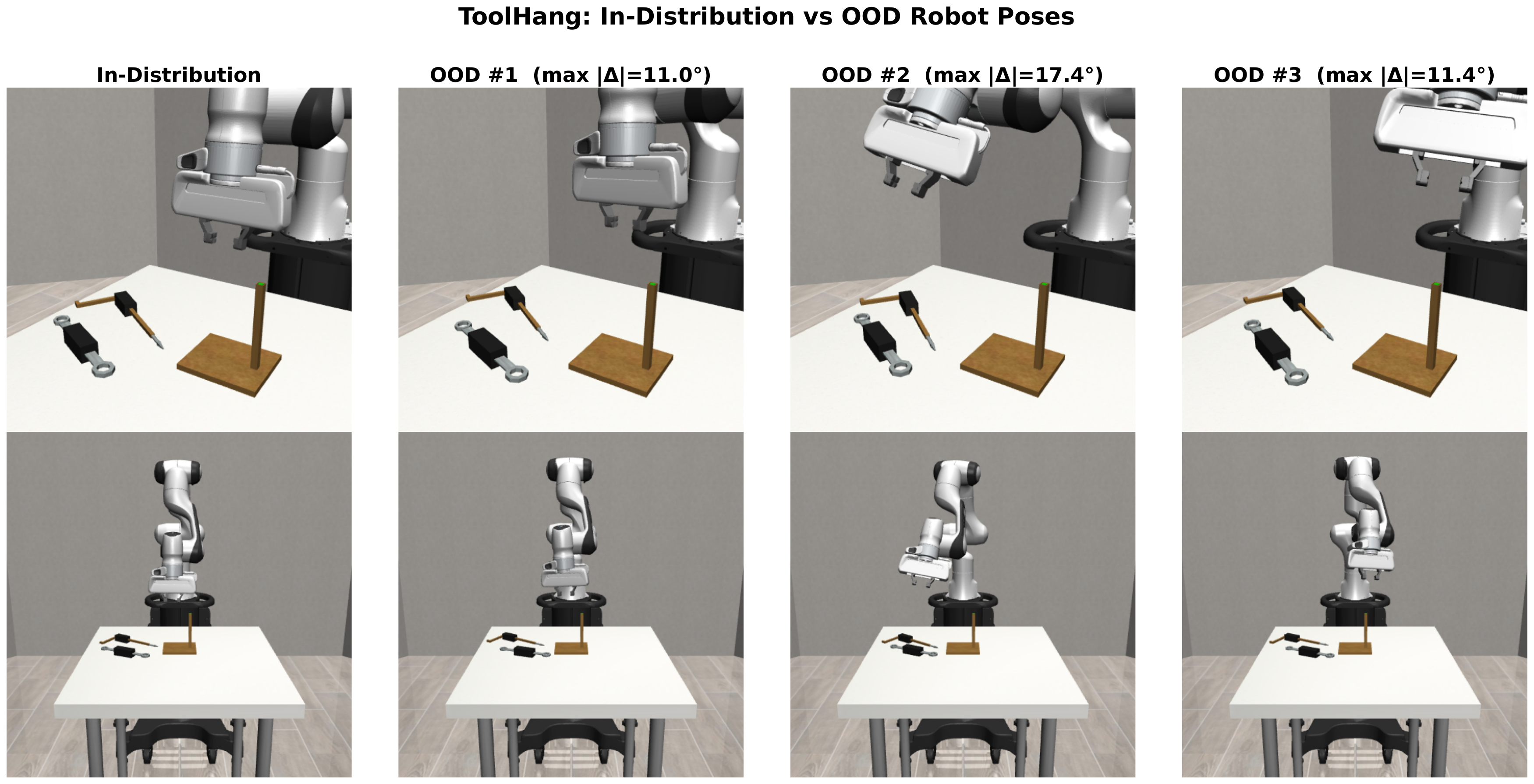}
    \caption{\textbf{ToolHang OOD initial states.} The layout follows
    Fig.~\ref{fig:robomimic-ood-square}. Top row: side view; bottom row:
    front view.}
    \label{fig:robomimic-ood-toolhang}
\end{figure}

We use $\sigma=0.1\,\mathrm{rad}$ in the main experiments. This perturbation is
large enough to move the robot away from the demonstration manifold, while
remaining physically feasible for task execution. The same perturbed initial
states are used for all baselines and our method to ensure a fair comparison.
Figs.~\ref{fig:robomimic-ood-square}, \ref{fig:robomimic-ood-transport},
and~\ref{fig:robomimic-ood-toolhang} visualize the resulting OOD initial states
for Square, Transport, and ToolHang. In each figure, the leftmost column shows
the in-distribution demonstration start, while the remaining columns show
independently sampled OOD starts. 

\section{LIBERO-Plus single-task settings}
\label{appendix: libero_plus_tasks}

We evaluate on four single-task settings from the LIBERO-Plus
benchmark, covering representative manipulation tasks
from the \emph{Kitchen} and \emph{Living Room} scenes of LIBERO-10. 
For robot-initial-state evaluation, the task instruction, scene layout,
camera configuration, lighting, and object categories match the corresponding
LIBERO-10 task. The distribution shift is introduced through perturbed
initial robot configurations following the official LIBERO-Plus
\emph{Robot Initial States} protocol. This setting allows us to evaluate
whether the policy and world model remain robust when the robot starts
from configurations outside the demonstration distribution, while the
task semantics remain unchanged.

\paragraph{Tasks.}
The four tasks used in our experiments are summarized in
Table~\ref{tab:libero_plus_tasks}. They include sequential stove
manipulation, microwave-door interaction, long-horizon dual-object
placement, and relational tabletop placement. These tasks cover both
single- and multi-stage manipulation, as well as rigid-object and
articulated-object interactions. We select these tasks because they cover diverse manipulation challenges while
keeping the distribution-shift factor controlled: Kitchen-S3 and Kitchen-S6
involve articulated-object interaction and sequential execution, Kitchen-S8
requires long-horizon placement of multiple objects, and LR-S6 tests relational
object placement in a different scene layout. Since the task instruction and
scene semantics remain unchanged, these settings isolate robustness to robot
initial-state shifts from changes in task specification or visual domain.

\paragraph{Demonstrations and evaluation protocol.}
For each task, we use the 50 expert demonstrations from the corresponding
LIBERO-10 task as the only supervision. No LIBERO-Plus demonstrations are
used during training. At evaluation time, we use the perturbed initial
states released by LIBERO-Plus under the \texttt{libero\_10} suite and the
\texttt{Robot Initial States} category. We evaluate 44 variants for Kitchen-S3, 42 for Kitchen-S6, 34 for Kitchen-S8, and 42 for LR-S6, following the
official protocol from LIBERO-Plus benchmark. We report the average success
rate over all evaluated initial-state variants.

\begin{table}[h]
\centering
\small
\caption{LIBERO-Plus single-task settings used in our experiments.
All tasks follow the official \texttt{Robot Initial States} protocol.}
\begin{tabular}{l l p{0.43\linewidth} c}
\toprule
\textbf{Alias} & \textbf{Scene} & \textbf{Language instruction} & \textbf{\# variants} \\
\midrule
Kitchen-S3 & KITCHEN\_SCENE3 
& turn on the stove and put the moka pot on it 
& 44 \\
Kitchen-S6 & KITCHEN\_SCENE6 
& put the yellow and white mug in the microwave and close it 
& 42 \\
Kitchen-S8 & KITCHEN\_SCENE8 
& put both moka pots on the stove 
& 34 \\
LR-S6 & LIVING\_ROOM\_SCENE6 
& put the white mug on the plate and put the chocolate pudding to the right of the plate 
& 42 \\
\bottomrule
\end{tabular}
\label{tab:libero_plus_tasks}
\end{table}

\section{Rollout data collection}
\label{appendix: rollout data collection}
We include rollout-data baselines to test whether the gains of feedback world model can be matched by simply collecting additional policy executions and retraining. To avoid information leakage, all rollout trajectories are collected in the original in-distribution Robomimic or LIBERO environments. None of the rollout datasets contains trajectories from the final robot-initial-state OOD evaluation setting.

\paragraph{Robomimic.}
For each Robomimic task, we first train the base diffusion policy using the same low-data setting as in the main experiments, i.e., 20\% of the official expert demonstrations. We then deploy intermediate policy checkpoints in the original Robomimic environment and record closed-loop trajectories. We collect 270 rollout episodes for Square and 150 rollout episodes for Transport and Tool-Hang. The rollout checkpoints are sampled across training to include both partially trained and stronger policies, producing a mixture of successful and failed behaviors. 

For data-augmentation baselines, we construct two behavior-cloning datasets: \emph{Mixed BC}, which combines the 20\% expert subset with all collected rollout trajectories, and \emph{Filtered BC}, which combines the same expert subset with only successful rollout trajectories. For the rollout-enhanced world model baseline, we train the world model on all official expert demonstrations plus all collected rollout trajectories, matching the setting used to evaluate whether additional offline data can improve prediction quality.

\paragraph{LIBERO-Plus single-task setting.}
For LIBERO-Plus, rollout data are collected from the four selected LIBERO-10 single tasks: Kitchen-S3, Kitchen-S6, Kitchen-S8, and LR-S6. For each task, rollout trajectories are generated by the corresponding single-task policy using the original LIBERO-10 task specification and language instruction. Episodes are initialized from the training demonstration initial states in a round-robin manner to keep the rollout distribution aligned with the expert-data distribution. We collect rollouts from multiple policy checkpoints with a maximum horizon of 500 environment steps. The resulting rollout sets contain 150 episodes for Kitchen-S3 and Kitchen-S6 respectively, 100 episodes for Kitchen-S8 and LR-S6 respectively.

As in Robomimic, we build two rollout-augmented policy-training datasets. \emph{Mixed BC} uses the original expert demonstrations together with all collected rollout episodes, regardless of success. \emph{Filtered BC} uses the same expert demonstrations but retains only rollout episodes that achieve task success. The rollout-enhanced world model is trained on the corresponding mixed rollout-augmented dataset. These baselines therefore represent a direct offline-data-augmentation alternative to our method, whereas the feedback world model uses no additional rollout data and instead corrects the world model state online from observations produced during the current inference episode.

\section{Complete inference pipeline}
\label{apendix: inference pipeline}

Algorithm~\ref{alg:fwm-inference} summarizes the complete inference
pipeline of our method. The procedure consists of an offline preprocessing stage
and an online deployment stage. The offline stage constructs the expert latent
memory and estimates the action-aware controllability weights, while the online
stage performs feedback world model guidance during diffusion-policy
sampling.

\paragraph{Offline preprocessing.}
Given the expert demonstration set $\mathcal{D}_{\rm expert}$, we first encode
all expert observations into the latent space of the observation encoder
$\psi$, yielding an expert latent memory
$\mathcal{Z}^{E}=\{\psi(O)\mid O\in\mathcal{D}_{\rm expert}\}$.
This memory is later used as a task-progress prior during guidance: predicted
future latents are encouraged to approach nearby expert latents. In addition,
we compute the action-aware weights $\{w_j^{(\beta)}\}_{j=1}^{D}$ from the
demonstrations via Eqs.~\eqref{eq:weight_calcu}--\eqref{eq:weight_normalize}. These weights estimate the degree to which each latent dimension
is controllable by the robot action, and are fixed throughout deployment. As a
result, the guidance objective focuses more on action-relevant latent changes
and suppresses the influence of latent components that are visually salient but
weakly controllable.

\paragraph{Online deployment.}
At the beginning of an episode, the robot observes $o_0$ and obtains its latent
state $z_0=\psi(o_0)$. We initialise the internal feedback state as
$\bar z_0=z_0$. At each environment timestep $t$, the diffusion policy samples
an initial noisy action sequence $A_t^{(T)}$ and iteratively denoises it from
$\tau=T$ to $\tau=1$. During high-noise denoising steps, we use the original
policy score $s_\phi(A_t^{(\tau)},O_t,\ell,\tau)$ without additional guidance.
Guidance is only applied in the final $\tau_g$ low-noise denoising steps, where
the action sequence already contains meaningful task structure and the
world model prediction is more informative.

For each guided denoising step, the feedback world model predicts the next
latent state under the current candidate action sequence. It evaluates the
base predictor at the observed latent $z_t$ and adds the correction
$\delta t L(z_t-\bar z_t)$. The feedback state $\bar z_t$ carries information
from previous interactions, allowing observed prediction residuals to
refine the candidate future prediction. The predicted
latent is then matched to its nearest expert latent $z^E_{i^\star}$ in
$\mathcal{Z}^{E}$, and the action-aware guidance energy
$E_{\rm ctrl}(A_t^{(\tau)})$ is computed using the controllability-weighted
latent distance via Eq.~\eqref{eq:Ectrl_calcu}. The resulting gradient is added to the diffusion-policy score,
producing the guided score
$\tilde{s}$ used for the next denoising update. This forms the inner guidance
loop of our method.

After denoising, the clean action sequence $A_t=A_t^{(0)}$ is obtained and the
first $T_a$ actions are executed in the environment. The feedback state is then
propagated using the corrected latent dynamics induced by the executed action:
\[
    \bar z_{t+1}
    =
    \bar z_t + \delta t \cdot \hat v_t(A_t^{(0)}).
\]
The next observation $o_{t+1}$ is encoded as $z_{t+1}=\psi(o_{t+1})$, and
the residual
\[
    e_{t+1}=z_{t+1}-\bar z_{t+1}
\]
is formed for next-step guidance. This residual measures the mismatch between
the feedback world model's internal belief after executing $A_t^{(0)}$ and the
actual latent state reached by the environment. It is then held fixed during the
next denoising process and used to correct all candidate world model
predictions at timestep $t+1$.

\paragraph{Role of the two key components.}
The feedback world model and action-aware guidance address two complementary
limitations of standard world model guided diffusion policies. The feedback
world model reduces prediction drift during inference by closing the loop
between predicted latent dynamics and actual observed transitions. This is
particularly important under robot initial-state-induced distribution shifts,
where open-loop predictions can quickly deviate from the true trajectory. In
contrast, the action-aware guidance improves how the corrected predictions are
used for policy guidance. Rather than treating every latent dimension equally,
it emphasizes dimensions that are more sensitive to robot actions and therefore
more useful for controlling task progress. Together, these two components yield
a closed-loop inference procedure in which the world model provides more
accurate future-state estimates, and the diffusion policy receives guidance
concentrated on action-controllable task-relevant changes.

\begin{algorithm}[!thb]
\caption{Inference Pipeline}
\label{alg:fwm-inference}
\begin{algorithmic}[1]
\Require Diffusion policy score $s_\phi$, encoder $\psi$, world model $f_\theta$, expert demos $\mathcal{D}_{\text{expert}}$, feedback gain $L$, controllability strength $\beta$, guidance window $\tau_g\!\le\!T$.

\vspace{2pt}
\Statex \textbf{Offline preprocessing:}
\State $\mathcal{Z}^{E}\!\gets\!\{\,\psi(O)\mid O\in\mathcal{D}_{\text{expert}}\,\}$;\quad compute action-aware weights $\{w_j^{(\beta)}\}_{j=1}^{D}$ via Eqs.~\eqref{eq:weight_calcu}--\eqref{eq:weight_normalize}.

\vspace{2pt}
\Statex \textbf{Online deployment:}
\State Observe $o_0$, encode $z_0=\psi(o_0)$, initialise feedback state $\bar z_0\!\gets\!z_0$.
\For{each timestep t}
    \State Sample $A_t^{(T)}\!\sim\!\mathcal{N}(0,\mathbf{I})$.
    \For{denoising step $\tau=T,\dots,1$}
        \If{$\tau\le\tau_g$} \Comment{guidance applied only at the last $\tau_g$ low-noise steps}
            \State Predict next latent with feedback-corrected dynamics: $z^{\mathrm{fb}}_{t+1}(A_t^{(\tau)})$ via Eq.~\eqref{eq:core-update-formula}.
            \State Retrieve nearest expert latent $z_{i^\star}^{E}$ (Eq.~\eqref{eq:nearest_expert}) and form $E_{\mathrm{ctrl}}(A_t^{(\tau)})$ (Eq.~\eqref{eq:Ectrl_calcu}).
            \State $\tilde s \gets s_\phi(A_t^{(\tau)},O_t,\ell,\tau) - \lambda\,\nabla_{A_t^{(\tau)}}E_{\mathrm{ctrl}}(A_t^{(\tau)})$. \Comment{Eq.~\eqref{eq:wm_guidance}}
        \Else
            \State $\tilde s \gets s_\phi(A_t^{(\tau)},O_t,\ell,\tau)$.
        \EndIf
        \State Denoise the action sequence from $A_t^{(\tau)}$ to $A_t^{(\tau-1)}$ using $\tilde{s}$.
    \EndFor
    \State Execute first $T_a$ actions of $A_t = A_t^{(0)}$ in $\mathcal{E}$.
\State Advance feedback state $\bar z_{t+1} \gets \bar z_t + \delta t\cdot \hat{v}(z_t, A_t^{(0)})$ via Eq.~\eqref{eq:zbar-update}.
\State Receive $o_{t+1}$, encode $z_{t+1} = \psi(o_{t+1})$. \Comment{form $e_{t+1}=z_{t+1}-\bar{z}_{t+1}$ via Eq.~\eqref{eq:residual error} for next-step guidance}

\EndFor
\end{algorithmic}
\end{algorithm}

\section{Implementation details.}
\label{appendix:implementation-details}

\paragraph{Lightweight latent world model architecture.}
The seven-task and real-world experiments use a lightweight, task-specific world model. The separate pretrained Fast-WAM experiment is described in Appendix~\ref{app:foundation}. For the lightweight models,
we use a compact latent dynamics model trained from the same task
demonstrations as the policy. The model predicts the next latent state from
the current observation latent, proprioceptive state, and the executed action
chunk. Each RGB view is encoded by the visual encoder associated with the
deployed diffusion policy, and the resulting visual features are fused with a
small proprioception encoder and an action encoder. A Transformer predictor
then maps this fused representation to the next-step latent state. The model
is trained with a one-step latent prediction objective and does not reconstruct
pixels or generate future videos.

This design keeps the world model small and data-efficient: it is trained per
task from the available demonstrations only, without large-scale pretraining
or additional real-world rollout data. During deployment, the world model is
frozen. Our feedback mechanism only maintains and corrects an online latent
state estimate using the latest real observation; it does not update the
world model parameters at test time.

\subsection{Implementation details of simulated tasks.}
\label{appendix:impl-sim}

\paragraph{Robomimic tasks.}
We evaluate three Robomimic tasks: Square, Transport, and Tool-Hang. For each
task, we train an image-conditioned diffusion policy under a low-data setting
using $20\%$ of the official proficient-human demonstrations. The policy uses
RGB observations and proprioception as input. Square and Tool-Hang are
single-arm tasks, while Transport is a dual-arm task. The visual observations
follow the standard task camera setup, using dual-view inputs such as the
agent/shoulder view and wrist-mounted camera view. The policies are trained
with a UNet-based diffusion policy, AdamW optimization, a cosine learning-rate
schedule, EMA, and DDPM denoising. Square uses a shorter action horizon, while
Transport and Tool-Hang use a longer action horizon to match their longer
execution sequences.

For each Robomimic task, we train a separate lightweight latent world model
using the same task demonstrations. The world model uses the same observation
modalities as the deployed policy. It predicts the next latent state after the
executed action chunk and is queried during diffusion-policy inference to
provide world model guidance. In the OOD evaluation, we perturb only the robot
initial joint configuration while keeping the task goal, object arrangement,
camera placement, and environment semantics unchanged. This produces a
robot-initial-state OOD setting without introducing visual domain
randomization.

\paragraph{LIBERO-Plus single-task tasks.}
We evaluate four LIBERO-10 single tasks under the Robot Initial States category
of LIBERO-Plus:
\texttt{KITCHEN\_SCENE3},
\texttt{KITCHEN\_SCENE6},
\texttt{KITCHEN\_SCENE8}, and
\texttt{LIVING\_ROOM\_SCENE6}. For each task, the policy and world model are
trained on demonstrations from the corresponding original LIBERO-10 task and
evaluated directly on the LIBERO-Plus robot-initial-state perturbation variants.

The LIBERO policies use dual-view RGB observations from the agent-view camera
and the eye-in-hand camera, together with proprioceptive state and the task
language embedding. We train a dedicated diffusion policy for each single task.
The corresponding lightweight world model is trained on the same task
demonstrations and uses the same visual, proprioceptive, action, and language
inputs as the policy. During evaluation, the world model predicts future
latent states under candidate action chunks, and the feedback mechanism
corrects its latent state estimate after each real environment observation.

\paragraph{Baselines and method configuration.}
We compare against the base diffusion policy, vanilla world model guidance,
and rollout-augmentation baselines when available. The base policy is executed
without world model guidance. Vanilla world model guidance uses the frozen
latent world model during diffusion denoising but does not use feedback
correction. Rollout-augmentation baselines retrain the diffusion policy with
additional policy rollouts: Mixed BC uses all collected rollouts, while
Filtered BC uses only successful rollouts. These rollouts are collected from
the original validation environments rather than from the final OOD evaluation
setting.

For our method, we use the same trained policy and the same trained base world
model as the vanilla guidance baseline. The only change is at inference time:
after each executed action chunk, the new observation is encoded and used to
correct the world model latent state before the next guidance step. When
action-aware weighting is enabled, we compute per-dimension
controllability weights from demonstration observations by measuring
counterfactual action variance in latent space offline.

\subsection{Implementation details of real-world tasks.}
\label{appendix:impl-real}

\paragraph{Robot platform.}
We conduct all real-world experiments on an R1 Lite robot. Although the
platform has two arms, all real-world experiments in this paper use only the
right arm and the right gripper; the left arm is kept idle throughout data
collection and evaluation. The policy outputs a right-arm control command and
a right-gripper command. The proprioceptive input contains the right-arm joint
state and the right-gripper state. To construct a real-world
robot-initial-state OOD setting, we reset the right arm to abnormal initial
poses outside the training distribution before evaluation, while keeping the
task goal and workspace layout unchanged.

\paragraph{Peach P\&P.}
Peach P\&P is a pick-and-place task in which the robot must grasp a peach from
the table and place it into a target container. We collect 50 expert
demonstrations on the R1 Lite platform using the right arm only. The final
deployed policy and world model use two camera views: an agent-view RGB camera
and the right-wrist RGB camera. The agent view provides global context of the
tabletop scene and the target container, while the right-wrist camera provides
close-range visual feedback during grasping and placement.

We train a task-specific image-conditioned diffusion policy from scratch on
the 50 demonstrations. The policy uses two observation frames and predicts an
action chunk for right-arm execution. RGB observations are resized and cropped
before being passed to the visual encoder. The lightweight world model is
trained from the same 50 demonstrations and uses the same two visual views,
right-arm proprioception, right-gripper state, and right-arm action input as
the policy. Thus, both the policy and world model are trained only from the
expert demonstration set, without additional real-world rollout data.

\paragraph{Drawer-Open.}
Drawer-Open is a more contact-rich manipulation task. The robot must locate
the drawer handle, align the gripper, establish contact, and pull the drawer
open with a suitable direction and force. The handle is made of deformable
rubber, so the task is sensitive to wrist pose, contact timing, and local
visual feedback near the handle. We collect 65 expert demonstrations using the
right arm of R1 Lite.

For Drawer-Open, the final deployed policy and world model use only the
right-wrist RGB camera view, together with right-arm and right-gripper
proprioception. We use this wrist-only observation setting because the task
depends primarily on local handle geometry, gripper-handle alignment, and
contact evolution after the wrist approaches the drawer. We train a
task-specific diffusion policy from scratch on the 65 demonstrations. The
corresponding lightweight world model is trained from the same demonstrations
and uses the same right-wrist visual observation and proprioceptive/action
inputs as the policy.

\paragraph{Real-world deployment protocol.}
At deployment time, the base diffusion policy generates action chunks from
the current observation history. For world-model-guided methods, the frozen
task-specific world model evaluates candidate denoising trajectories in latent
space. After each executed action chunk, the latest real observation from the
robot is encoded and used as feedback to correct the world model latent state
before the next guidance step. This online correction is applied only at
inference time and does not update the policy or world model parameters. All real-world methods are evaluated under the same reset protocol and task
success criteria. Peach P\&P is counted as successful when the peach is
grasped and placed into the target container. Drawer-Open is counted as
successful when the drawer is pulled open to the required extent.

\subsection{Compute resources.} 
\label{sec:compute resources}
For the task-specific policy and world-model experiments under robot
initial-state shifts, training and evaluation run on an internal GPU
workstation/cluster with one RTX A6000 GPU (48\,GB memory),
8--16 CPU cores, and 64--128\,GB host memory per job.
These experiments require approximately 400--700 single-GPU hours,
including policy training, world-model training, hyperparameter sweeps,
and evaluation. Preliminary and exploratory runs used additional compute,
roughly $1.5$--$2\times$ this budget. Inference cost measured on an
RTX PRO 6000 is reported in Appendix~\ref{app:latency}.

\subsection{Existing assets and licenses}
\label{appendix:assets-licenses}

We use public robotic benchmarks and assets only for research purposes. Robomimic is cited in the main paper and used following its public release and license. LIBERO and LIBERO-Plus are cited in the main paper and used following their official benchmark instructions. At the time of writing, the official LIBERO-Plus repository does not provide an explicit license file. We do not redistribute or repackage these assets in this work. All real-world demonstrations are collected on our own robot platform.


\MarsLabAppendixLayout
\section{Evaluation statistics}
\label{app:statistics}
\paragraph{Prediction error.}
For Fig.~\ref{fig:wm_mse}, we evaluate latent MSE one executed action
chunk ahead. Simulation MSE is
averaged over all valid prediction steps, with standard errors estimated
by resampling complete episodes 10,000 times. Real-world MSE uses the
mean of episode means and its SEM. These error bars quantify variability
across evaluation episodes for fixed trained models.
Table~\ref{tab:wm_mse_statistics} lists the values and episode counts.

\begin{table}[htbp]
\centering\small
\caption{Latent prediction MSE in Fig.~\ref{fig:wm_mse}, scaled by $10^3$.
Simulation values are step-pooled mean $\pm$ episode-cluster bootstrap SE;
real-world values are episode-equal mean $\pm$ SEM. $n$ is the number of
episodes per available method.
Real-world rollout models were not evaluated.}
\label{tab:wm_mse_statistics}
\begin{tabular}{lrccc}
\toprule
Task & $n$ & Base WM & Rollout-Enhanced WM & Feedback WM\\
\midrule
Square & 50 & $16.246\pm3.904$ & $15.472\pm1.092$ & $4.333\pm0.528$\\
Tool-Hang & 50 & $8.399\pm1.006$ & $10.295\pm1.647$ & $5.373\pm0.545$\\
Transport & 50 & $16.248\pm2.617$ & $13.160\pm1.832$ & $10.432\pm0.495$\\
Kitchen-S3 & 44 & $15.676\pm0.820$ & $17.510\pm0.740$ & $13.400\pm0.633$\\
Kitchen-S6 & 42 & $11.999\pm0.920$ & $11.698\pm0.829$ & $8.959\pm0.484$\\
Kitchen-S8 & 34 & $10.189\pm1.073$ & $9.514\pm0.908$ & $8.553\pm0.680$\\
LivingRoom-S6 & 42 & $20.212\pm0.907$ & $18.044\pm0.582$ & $15.831\pm0.726$\\
Peach-P\&P & 20 & $2.177\pm0.169$ & n/a & $0.754\pm0.107$\\
Drawer-Open & 20 & $1.309\pm0.099$ & n/a & $0.309\pm0.080$\\
\bottomrule
\end{tabular}
\end{table}

\section{Evaluation under LIBERO-Plus perturbations}
\label{app:broader_ood}
We evaluate four LIBERO-Plus tasks under five perturbation categories:
robot initial states, lighting, background textures, object layout,
and sensor noise. Each category is evaluated on the same four tasks,
yielding 20 task--perturbation settings. Table~\ref{tab:ood_full} reports
success rates for each setting, and Table~\ref{tab:broader_ood} summarizes
the results pooled within each category.

\begin{table}[htbp]
\centering\small
\caption{Success rates for all five LIBERO-Plus perturbation categories.
K-S3, K-S6, K-S8, and LR-S6 denote the four tasks in
Appendix~\ref{appendix: libero_plus_tasks}.}
\label{tab:ood_full}
\begin{tabular}{llrrrr}
\toprule
Perturbation & Method & K-S3 & K-S6 & K-S8 & LR-S6\\
\midrule
Robot initial states & Base & 0.55 & 0.45 & 0.59 & 0.26\\
 & Ours & 0.68 & 0.55 & 0.65 & 0.36\\
Lighting & Base & 0.50 & 0.21 & 0.42 & 0.58\\
 & Ours & 0.40 & 0.24 & 0.58 & 0.74\\
Background textures & Base & 0.97 & 0.50 & 0.55 & 0.38\\
 & Ours & 0.94 & 0.68 & 0.58 & 0.44\\
Object layout & Base & 0.81 & 0.74 & 0.48 & 0.63\\
 & Ours & 0.85 & 0.85 & 0.64 & 0.66\\
Sensor noise & Base & 0.98 & 0.88 & 0.80 & 0.59\\
 & Ours & 0.98 & 0.90 & 0.87 & 0.70\\
\bottomrule
\end{tabular}
\end{table}

Our method matches or exceeds the base policy in 18 of the 20
task--perturbation settings (Table~\ref{tab:ood_full}). Pooled success
increases in all five categories, by 4.5--9.9 percentage points, alongside
reductions in latent MSE (Table~\ref{tab:broader_ood}). For example,
success on Kitchen-S6 under background changes rises from 50\% to 68\%,
while Kitchen-S8 under object-layout changes improves from 48\% to 64\%.
These results demonstrate improvements in both aggregate prediction
accuracy and policy performance across the evaluated distribution shifts.

\section{Action dependence and physical-state validation}
\label{app:validation}
\subsection{Action-dependent prediction}
Effective world-model guidance requires predictions that respond to the
chosen action. Table~\ref{tab:action_controls} shows that using correctly
paired actions yields substantially lower latent MSE than persistence,
a constant training-mean prediction, or shuffled actions. Shuffling the
actions while keeping the predictor fixed increases MSE by approximately
$20.9\times$ on Square and $36.6\times$ on Transport, indicating that action
information contributes substantially to predictive accuracy. The advantage
over persistence and the training mean further supports the model's ability
to capture action-conditioned transitions beyond state continuity and
average latent structure.

\begin{table}[htbp]
\centering\small
\caption{Action-conditioning controls. Values are latent MSE.}
\label{tab:action_controls}
\begin{tabular}{lrr}
\toprule
Predictor & Square & Transport\\
\midrule
Persistence & 0.0190 & 0.0324\\
Constant training mean & 0.0382 & 0.0446\\
WM with shuffled actions & 0.0335 & 0.0914\\
WM with correct actions & 0.0016 & 0.0025\\
\bottomrule
\end{tabular}
\end{table}

\begin{figure}[htbp]
\centering
\includegraphics[width=2.7in]{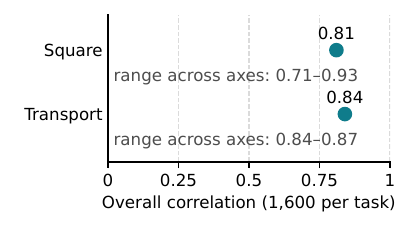}
\caption{\textbf{Controlled action effects.} Overall correlation between
predicted and realized end-effector effects over 1,600 interventions per
task, from 200 saved states with eight action branches each. Text labels
give the range of per-axis correlations.}
\label{fig:action_interventions}
\end{figure}

Controlled interventions further connect this action dependence to physical
motion. We vary the action from identical OOD simulator states and use a
frozen probe to decode the predicted end-effector effects. Predicted and
realized effects have overall correlations of 0.81 on Square and 0.84 on
Transport (Fig.~\ref{fig:action_interventions}), with per-axis correlations
of 0.71--0.93 and 0.84--0.87, respectively. This agreement suggests that the
model captures how action changes affect robot motion. Together, the two
comparisons support using its predictions to distinguish the physical
consequences of candidate actions for policy guidance.

\subsection{Physical information in the latent representation}
Physical-state probes show that the learned latent representation retains
information about robot configuration and object position. We fit
episode-disjoint ridge probes to the frozen task-trained latent, with
ImageNet-pretrained ResNet-18 features from the same observations as a
reference (Table~\ref{tab:physical_probes}). The recovered end-effector
positions and orientations, together with object and payload positions,
connect the latent representation to physical quantities relevant to
manipulation.

\begin{table}[htbp]
\centering\small
\caption{Physical-state probe RMSE. Slash-separated entries refer to
Transport arms 0/1.}
\label{tab:physical_probes}
\begin{tabular}{llcc}
\toprule
Task & Target & Task-trained & ImageNet control\\
\midrule
Square & EEF position (mm) & 34 & 29\\
 & EEF orientation (degrees) & 10.5 & 10.4\\
 & Object position (mm) & 30 & 30\\
Transport & EEF position (mm) & 87/57 & 77/61\\
 & EEF orientation (degrees) & 5.8/4.0 & 6.4/5.3\\
 & Payload position (mm) & 84 & 88\\
\bottomrule
\end{tabular}
\end{table}

The predicted latents preserve this physical information. Using the frozen
probes to decode world-model predictions gives position errors comparable
to the probes' held-out errors on observed latents
(Table~\ref{tab:physical_prediction}). For example, the decoded Square
end-effector and object positions have errors of 25.8\,mm and 21.9\,mm,
respectively, against probe errors of 29.6\,mm and 25.3\,mm. These results
indicate that the predictions capture physical state at a precision
consistent with the probe's decoding accuracy, complementing latent MSE
with an interpretable measure of robot and object position.

\begin{table}[htbp]
\centering\small
\caption{Physical error decoded from predicted latents and probe
resolution on observed latents. All values are in millimeters.}
\label{tab:physical_prediction}
\begin{tabular}{lrr}
\toprule
Target & Prediction error & Probe resolution\\
\midrule
Square EEF position & 25.8 & 29.6\\
Square object position & 21.9 & 25.3\\
Transport arm 0 EEF position & 55.9 & 64.6\\
\bottomrule
\end{tabular}
\end{table}

\subsection{Qualitative latent trajectories}
\begin{figure}[htbp]
\centering
\includegraphics[width=\linewidth]{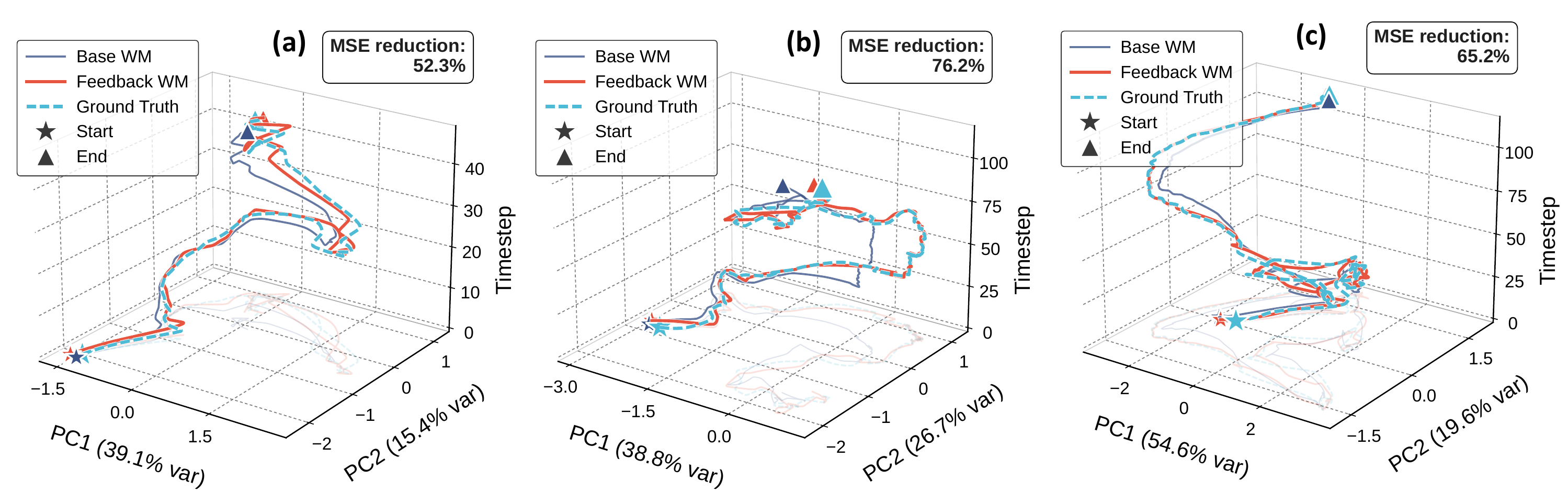}
\caption{\textbf{Online latent-trajectory tracking with feedback.}
Base WM, Feedback WM, and observed (ground-truth) latent trajectories
for three example episodes, shown in a shared PCA projection with time
as the third axis. Feedback correction brings predictions closer to the
observed trajectories through changes in direction and across execution
stages. Stars and triangles mark the start and end of each trajectory.}
\label{fig:trajectory}
\end{figure}

To examine how feedback correction evolves during execution,
Fig.~\ref{fig:trajectory} compares the base and feedback world-model
predictions with the observed latent trajectories in a shared PCA
projection. The time axis reveals when prediction errors emerge and how
subsequent feedback updates affect trajectory tracking.

In cases (a) and (b), the base model develops noticeable departures from
the observed trajectory during the middle and later stages of execution.
The feedback-corrected predictions follow the observed bends and changes
in direction more closely, reducing these persistent offsets. Case (c)
highlights the correction of early-stage deviations, followed by close
tracking as execution proceeds. These examples show how online observation
feedback helps the predictor follow the evolving state across different
phases of an episode. They complement the aggregate MSE results in
Fig.~\ref{fig:wm_mse} by illustrating the temporal behavior of the
correction: prediction discrepancies observed during execution are used
to improve alignment in subsequent predictions, without updating model
parameters.

\section{Transfer to Fast-WAM-Joint}
\label{app:foundation}
\paragraph{Model and feedback implementation.}
We implement feedback on Fast-WAM-Joint~\citep{fast-wam}, which uses
a pretrained Wan2.2-5B backbone and two-camera inputs across
LIBERO-10. The architecture and learned parameters remain frozen.
Feedback operates in the video branch's native future-latent space.
For a predicted future latent $F$ and its observed counterpart $G$,
encoded by the same VAE, we maintain an exponential moving average
$b\leftarrow0.7b+0.3(G-F)$ and correct subsequent predictions as
$F_{\rm fb}=F+\kappa b$. The residual is incorporated only after the
corresponding observations become available, allowing past prediction
errors to inform subsequent predictions with the model parameters fixed.

\paragraph{Prediction evaluation.}
Each prediction covers four future frames at offsets of 8, 16, 24 and
32 environment steps. We compare base and corrected predictions on the
same recorded rollouts, without changing executed actions, and average
errors over paired prediction calls. The ID set contains 1,203 calls
from 100 episodes; the robot-initial-state OOD set contains 7,553 calls
from 393 episodes covering all 393 LIBERO-Plus variants across LIBERO-10.
Latent MSE is the mean squared difference in the native future-latent
coordinates. For pixel metrics, both predictions and the VAE-encoded
observations are decoded with the same VAE, including the initial frame
as causal context. We compute MSE and PSNR on the four future frames
at pixel range 0--255; the reference is the VAE reconstruction of the
observations.

Feedback improves latent prediction accuracy in both ID and OOD settings
at all three tested gains, 0.2, 0.3 and 0.5
(Fig.~\ref{fig:foundation_gain}). Within this range, increasing the gain
progressively reduces error. At gain 0.5, OOD latent MSE falls from 0.3015
to 0.2464, an 18.3\% reduction, compared with a 7.8\% reduction under ID
conditions. The larger relative improvement under distribution shift
suggests that observation feedback is particularly useful when the
pretrained predictor encounters a mismatch with deployment conditions.
These gains show that a large pretrained model can still benefit from
correcting its predictions using observations from the current execution.

The improvement also carries through to decoded visual predictions.
At gain 0.5, OOD pixel MSE decreases by 18.0\% and mean PSNR increases
by 0.86\,dB, with positive PSNR gains at all four prediction offsets
(0.76--0.94\,dB). The agreement between latent-space and image-space
metrics indicates that the correction improves the fidelity of the
predicted visual outcomes across the evaluated horizons. Because the
backbone remains frozen, these results demonstrate that feedback can
complement pretrained dynamics through inference-time correction alone.

\paragraph{Policy evaluation.}
To assess the utility of feedback for control, we evaluate policy success
on the LIBERO-Plus Robot Initial States OOD variants across LIBERO-10.
Our method improves Fast-WAM-Joint success from 77.1\% to 81.4\%, a gain
of 4.3 percentage points. Together with the prediction results, this
improvement supports the plug-and-play applicability of feedback to a
pretrained world-action model: the same model benefits in both prediction
accuracy and task performance while retaining its learned parameters.

\section{Comparison with filtering baselines}
\label{app:filtering}
Observation feedback is also central to classical filtering and learned
belief updates. We therefore compare these approaches to examine whether
our lightweight residual correction can improve OOD prediction while
reusing a frozen predictor. The comparison uses the same frozen encoder
and 50 expert demonstrations per task. An estimated-gain Kalman filter
(Est.\ KF) provides a classical filtering baseline, while a PlaNet-style
RSSM represents learned state estimation. RSSM Open predicts through its
learned prior; RSSM Filter incorporates observations through its posterior.

Table~\ref{tab:filtering_sem} shows that our method achieves the lowest
OOD latent MSE on all three tasks, reducing error by 31.2--57.1\% relative
to Base WM and by 8.4--21.6\% relative to Est.\ KF. The Kalman baseline
already improves on Base WM, demonstrating the value of observation-based
correction. The further reduction achieved by our method shows that its
residual-feedback mechanism provides an effective way to correct model
mismatch beyond this filtering baseline.

Within the RSSM, posterior conditioning reduces error relative to the
prior, confirming that observations help refine learned state estimates.
Our method achieves lower error by correcting the existing predictor
directly. In this shared-encoder, limited-data setting, the results favor
reusing learned dynamics with online residual correction: the method
exploits deployment observations without learning a posterior network or
updating model parameters. This offers a practical route to improved OOD
prediction when a trained world model is already available.

\begin{table}[htbp]
\centering\small
\caption{OOD latent MSE, mean $\pm$ SEM across stored trajectories.
The unit of replication is one trajectory, with $n=150$, $150$, and $95$
for the three columns.}
\label{tab:filtering_sem}
\begin{tabular}{lccc}
\toprule
Method & Kitchen-S3 & Kitchen-S6 & LivingRoom-S6\\
\midrule
Base WM & $0.04565\pm0.00040$ & $0.04250\pm0.00031$ & $0.04535\pm0.00044$\\
Est.\ KF & $0.03433\pm0.00026$ & $0.02325\pm0.00013$ & $0.03021\pm0.00032$\\
RSSM Open & $0.05895\pm0.00075$ & $0.05377\pm0.00036$ & $0.05630\pm0.00059$\\
RSSM Filter & $0.05698\pm0.00074$ & $0.04938\pm0.00027$ & $0.05313\pm0.00054$\\
Ours & $\mathbf{0.03143}\pm0.00036$ & $\mathbf{0.01823}\pm0.00015$ & $\mathbf{0.02671}\pm0.00038$\\
\bottomrule
\end{tabular}
\end{table}

\section{Empirical behavior of the dynamics residual}
\label{app:residual}
Feedback correction uses observed prediction errors to refine subsequent
predictions. To understand the model mismatch that this mechanism must
track, we examine how the base world model's residual evolves throughout
execution. The analysis asks how its magnitude changes across deployment
conditions and across stages of a task, complementing the aggregate
prediction errors in Fig.~\ref{fig:wm_mse}.

\begin{figure}[htbp]
\centering
\includegraphics[width=\linewidth]{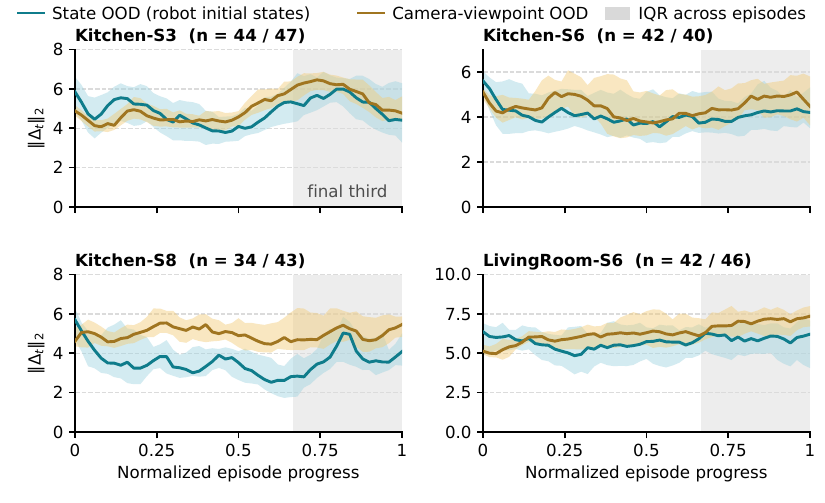}
\caption{\textbf{Temporal structure of the base world model's residual.}
Episodes are aligned from start to finish on a normalized progress axis.
Curves show the across-episode median; shaded bands span the 25th--75th
percentiles. Teal and brown denote robot-initial-state and camera-viewpoint
OOD, respectively. Gray marks the final third used in
Table~\ref{tab:residual_temporal}.}
\label{fig:residual_histories}
\end{figure}

\paragraph{Evaluation.}
We evaluate four LIBERO tasks under robot-initial-state and camera-viewpoint
shifts, keeping the task-specific policy, world model, and feedback gain
fixed across the two conditions. The residual magnitude is
\begin{equation}
\|\Delta_t\|_2
=\frac{\|z_{t+1}-f_\theta(z_t,A_t)\|_2}{\delta t},
\end{equation}
where $z_{t+1}$ is the observed next state in the weighted latent
representation used for feedback, $f_\theta(z_t,A_t)$ is the base model's
prediction, and $\delta t=0.75$\,s is the interval between updates.
Fig.~\ref{fig:residual_histories} shows the residual histories aligned by
episode progress. Table~\ref{tab:residual_temporal} summarizes their level
and relative variability during the final third of each episode.

\paragraph{Temporal patterns and implications.}
Fig.~\ref{fig:residual_histories} shows that residuals continue to vary
throughout execution. In Kitchen-S3, both conditions exhibit a lower
residual level around the middle of the episode followed by a later rise.
In Kitchen-S8, camera-viewpoint shifts produce higher residuals through
most of the episode, while the camera-shift residual in LivingRoom-S6
increases gradually with progress. These patterns reveal both temporal
structure within an execution and differences between deployment conditions.

The late-episode statistics in Table~\ref{tab:residual_temporal} reinforce
this observation: camera-viewpoint shifts yield higher residual levels
in all four tasks, including an increase from 3.56 to 4.89 on Kitchen-S8.
The coefficient of variation ranges from 0.11 to 0.21 across conditions,
indicating continued fluctuations around these levels. Together, the
curves and summaries motivate maintaining a feedback state throughout
execution: updating it with each new observation allows the correction
to respond to the current mismatch as the task progresses. This provides
an empirical rationale for the continuous observation--prediction loop
used by our method.

\paragraph{Connection to the theoretical analysis.}
The persistent residual levels and temporal variation in
Fig.~\ref{fig:residual_histories} motivate two complementary roles of
feedback: compensating persistent model mismatch and updating the
correction as execution progresses. Under the residual assumptions in
Eq.~\eqref{eq:bounded-residual-proof} and a stable gain, the observer
correction satisfies $Le_t\to\Delta_\infty$, cancelling the limiting
prediction bias (Eq.~\eqref{eq:corrected-prediction-error-limit-proof}).
This bias-compensation mechanism provides a theoretical interpretation
of the prediction improvements in Fig.~\ref{fig:wm_mse}. Incorporating
successive observations keeps the correction responsive to the evolving
mismatch, connecting the observer formulation to the online adaptation
that makes feedback useful under distribution shift.

\begin{table}[htbp]
\centering\small
\caption{\textbf{Late-episode residual statistics.}
$n$ is the number of episodes. Late residual is the median across episodes
of each episode's median residual in its final third. CoV is the median
across episodes of the within-window standard deviation divided by the
mean. State and Camera denote robot-initial-state and camera-viewpoint
OOD, respectively.}
\label{tab:residual_temporal}
\begin{tabular}{llrrr}
\toprule
Scene & Condition & $n$ & Late residual & CoV\\
\midrule
Kitchen-S3 & State & 44 & 5.17 & 0.18\\
 & Camera & 47 & 5.72 & 0.16\\
Kitchen-S6 & State & 42 & 4.17 & 0.15\\
 & Camera & 40 & 4.59 & 0.17\\
Kitchen-S8 & State & 34 & 3.56 & 0.21\\
 & Camera & 43 & 4.89 & 0.14\\
LivingRoom-S6 & State & 42 & 6.16 & 0.12\\
 & Camera & 46 & 6.98 & 0.11\\
\bottomrule
\end{tabular}
\end{table}

\section{Feedback gain and action-weight interpolation}
\label{app:sensitivity}
We study sensitivity to two complementary choices: the feedback gain $l$,
which controls prediction correction, and the interpolation coefficient
$\beta$, which controls the emphasis on action-responsive latent dimensions.
The experiments assess the range of settings that yield useful correction
and guidance, and inform parameter selection for deployment.

\paragraph{Feedback gain.}
For $L=lI$, the effective correction strength per update is $\delta t l$;
the LIBERO experiments use $\delta t=0.75$\,s.
Table~\ref{tab:gain_sweep} evaluates how changing $l$ affects latent
prediction accuracy on four tasks. Every tested gain from 0.02 to 0.5
reduces MSE relative to Base WM, with reductions of 7.6--24.9\%.
The best tested gains lie between 0.1 and 0.3, showing that substantial
improvements are available across several moderate settings. The results
at $l=1.0$ further favor moderate correction strengths in this evaluation.
Together, these observations support selecting the gain using held-out
prediction error within a practical operating range.

\begin{table}[htbp]
\centering\small
\caption{Latent MSE reduction relative to Base WM under gain variation.
Negative values denote degradation.}
\label{tab:gain_sweep}
\begin{tabular}{crrrr}
\toprule
$l$ & Kitchen-S3 & Kitchen-S6 & Kitchen-S8 & LR-S6\\
\midrule
0.02 & 7.6\% & 11.0\% & 10.8\% & 10.7\%\\
0.1 & 13.5\% & 23.0\% & 21.3\% & 19.9\%\\
0.2 & 16.0\% & 24.9\% & 18.1\% & 22.3\%\\
0.3 & 17.6\% & 24.3\% & 20.5\% & 22.7\%\\
0.5 & 9.5\% & 19.9\% & 12.8\% & 18.8\%\\
1.0 & $-20.0\%$ & $-9.4\%$ & $-12.5\%$ & $-8.6\%$\\
\bottomrule
\end{tabular}
\end{table}

The task gains used in our evaluations are $0.5/0.1/0.5$ for
Square/Tool-Hang/Transport and $0.05/0.5/0.05/0.1$ for
Kitchen-S3/Kitchen-S6/Kitchen-S8/LR-S6, selected using held-out latent MSE.

\paragraph{Action-weight interpolation.}
Table~\ref{tab:beta_sweep} examines how the balance between uniform and
action-aware weighting affects policy success, with other settings fixed.
Intermediate interpolation gives the highest tested success on both tasks:
Kitchen-S8 reaches 0.667 at $\beta=0.3$, and Kitchen-S3 reaches 0.697 at
$\beta=0.7$. These improve on full weighting ($\beta=1$) by 4.0 and
3.8 percentage points, respectively. This pattern supports the interpolation
in Eq.~\eqref{eq:weight_normalize}: retaining a uniform component while
emphasizing action-responsive dimensions can improve guidance. The
different maximizing values also indicate that the appropriate emphasis
can be adapted to the task. Across the tested coefficients, success varies
by only 4.0 and 3.8 percentage points on the two tasks, respectively.

\begin{table}[htbp]
\centering\small
\caption{Success rate under action-weight interpolation $\beta$ in
Eq.~\eqref{eq:weight_normalize}, with other settings fixed.}
\label{tab:beta_sweep}
\begin{tabular}{lrrrr}
\toprule
Task & $\beta=0.2$ & $\beta=0.3$ & $\beta=0.7$ & $\beta=1.0$\\
\midrule
Kitchen-S8 & 0.656 & 0.667 & 0.637 & 0.627\\
Kitchen-S3 & 0.667 & 0.682 & 0.697 & 0.659\\
\bottomrule
\end{tabular}
\end{table}

\section{Inference cost}
\label{app:latency}
\begin{figure}[htbp]
\centering
\includegraphics[width=\linewidth]{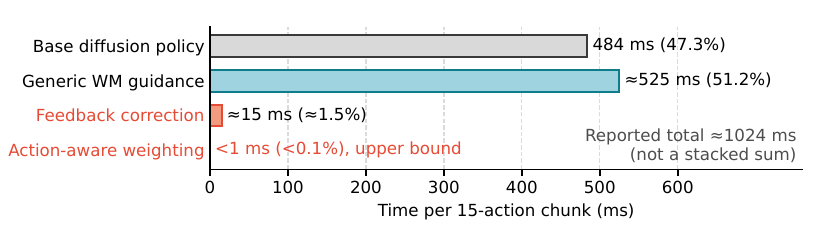}
\caption{\textbf{Inference cost by component.} Approximate time per
15-action chunk on one RTX PRO 6000, batch size 1, using DDPM-100 with
guidance in the final 15 denoising steps. Feedback correction and
action-aware weighting together account for less than 2\% of the total.}
\label{fig:latency_breakdown}
\end{figure}

We evaluate how much inference overhead feedback correction and
action-aware weighting introduce when integrated into world-model-guided
diffusion inference. To quantify their contribution, we measure the time
spent in each component (Fig.~\ref{fig:latency_breakdown}). Per 15-action
chunk, the base diffusion policy takes approximately 484\,ms, and generic
world-model guidance adds about 525\,ms. Feedback correction takes only
about 15\,ms, while action-aware weighting adds less than 1\,ms. Together,
our two modules account for less than 2\% of the approximately 1.02\,s
total inference time. Thus, the additional computation is dominated by
the underlying guidance procedure. Feedback correction and action-aware
weighting enhance this procedure with minimal marginal inference overhead.

\end{document}